\documentclass{article}
\usepackage{log_2025}						

\usepackage{booktabs}						
\usepackage{multirow}						
\usepackage{amsfonts}						
\usepackage{graphicx}						
\usepackage{duckuments}						
\usepackage{url}
\usepackage[misc]{ifsym}
\usepackage{nicefrac}       
\usepackage{microtype}      
\usepackage[table]{xcolor}
\usepackage{arydshln}
\usepackage{microtype}
\usepackage{graphicx}
\usepackage{subfigure}
\usepackage{dblfloatfix}
\usepackage{placeins}
\usepackage{multirow}		
\usepackage{amsmath}
\usepackage{amssymb}
\usepackage{mathtools}
\usepackage{etoolbox}
\usepackage{wrapfig}
\usepackage{makecell}
\usepackage{arydshln}
\usepackage{adjustbox}

\definecolor{firstcolor}{HTML}{009E73}
\definecolor{secondcolor}{HTML}{0072B2}
\definecolor{thirdcolor}{HTML}{D55E00}

\usepackage[numbers,compress,sort]{natbib}	

\title[Link Prediction with Untrained Message Passing Layers]{Link Prediction with Untrained Message Passing Layers}

\author[Qarkaxhija et al.]{%
Lisi Qarkaxhija\thanks{Equal contribution.} \quad Anatol E. Wegner\footnotemark[1] \quad Ingo Scholtes\\
Chair of Machine Learning for Complex Networks \\
Center for Artificial Intelligence and Data Science (CAIDAS) \\
Julius-Maximilians-Universität Würzburg, DE \\
\email{name.surname@uni-wuerzburg.de}
}

\begin{document}

\maketitle

\begin{abstract}
In this work, we explore the use of untrained message passing layers in graph neural networks for link prediction. The untrained message passing layers we consider are derived from widely used graph neural network architectures by removing trainable parameters and nonlinearities in their respective message passing layers. Experimentally, we find that untrained message passing layers can lead to competitive and even superior link prediction performance compared to fully trained message passing layers while being more efficient, especially in the presence of high-dimensional features. We also provide a theoretical analysis of untrained message passing layers in the context of link prediction and show that the inner product of features produced by untrained message passing layers relate to common neighbour and path-based topological measures which are widely used for link prediction. As such, untrained message passing layers offer a more efficient alternative to trained message passing layers in link prediction tasks, with clearer theoretical links to classical path-based heuristics.
\end{abstract}

\section{Introduction}
\label{section: Introduction}

Graph neural networks (GNNs) are a powerful class of machine learning models that can learn from graph-structured data, such as social networks, molecular graphs, and knowledge graphs. 
GNNs have emerged as an important tool in the machine learning landscape, due to their ability to model complex relationships and dependencies within data and have found applications in a variety of fields where data exhibits a complex topology that can be captured as a graph. 
This is shown by a multitude of studies, including \cite{basu2022equivariant, wang2022approximately, fu2022simulate, jaini2021learning, puny2021frame}, which highlight the versatility and adaptability of GNNs for machine learning tasks across a range of fields.

One of the key concepts underlying GNNs is Message Passing (MP) \citep{gilmer2017neural}, which operates by propagating and aggregating information between nodes in the graph, using message and update functions possibly with learnable parameters. However, designing effective GNNs can be challenging, as they can suffer from issues such as over-smoothing or over-parameterization, and because training GNNs can be computationally demanding. In order to address these shortcomings recent efforts have concentrated on finding simplified architectures that are both more interpretable and easier to optimize.

With our work, we aim to complement existing works on simplified and untrained MP architectures, previously formulated in the context of node classification, from the perspective of link prediction \citep{zhang2018link}. Link prediction (LP) is an important task in graph ML with many applications such as recommender systems, spam mail detection, drug re-purposing, and many more \citep{zhang2018link}. Current state-of-the-art LP methods \citep{wang2023neural} in general rely on a combination of GNNs and structural features; hence formulating effective and efficient MP architectures has the potential to further improve current LP methods in these regards. Therefore, we focus our analysis on MP layers rather than trying to formulate a novel end-to-end LP method.   

In this work, we explore the use of untrained message passing layers for link prediction in graph datasets with high dimensional features.
By formulating Untrained Message Passing (UTMP) layers, we follow an approach similar to that of \emph{Simplified Graph Convolutional Networks} introduced by \cite{wu2019simplifying}. 
This approach simplifies GNN architectures by removing trainable parameters and nonlinearities resulting in an architecture that can be clearly separated into two components: an untrained message passing/feature propagation steps followed by a linear classifier. In addition to these we also consider fully untrained architectures based on simple inner products of features obtained after $l$ iterations of UTMP layers as a baseline and find that these features produced by UTMP layers are already highly informative leading to surprisingly high link prediction performances.  

We base our analysis on untrained versions of four widely used MP architectures, namely Graph Convolutional Networks (GCN)\citep{kipf2016semi}, SAGE \citep{hamilton2017inductive}, GraphConv \citep{morris2019weisfeiler} and GIN \citep{xu2018powerful}. 
We test these untrained message passing layers on a variety of datasets that cover a wide range of sizes, node features, and topological characteristics ensuring a comprehensive evaluation of the models. 
We test UTMP layers on a variety of datasets that cover a wide range of sizes, node features, and topological characteristics ensuring a comprehensive evaluation of the UTMP layers. Mirroring the results reported by \cite{wu2019simplifying} for node classification we find that UTMP layers in many cases outperform their fully trained counterparts in LP tasks, while being easier to optimize and exhibiting clearer theoretical connections to classical path-based measures.

We also show that link prediction provides a complementary perspective for the theoretical analysis of UTMP layers \citep{zhou2022ood, chamberlain2022graph}. In our theoretical analysis we establish a direct connection between features produced by UTMP layers and various path based node similarity measures. Path based measures capture the indirect connection strength between node pairs nodes in the absence of a direct link connecting the nodes. Consequently, path based measures and methods have been widely used in traditional link prediction methods \citep{martinez2016survey, kumar2020link} and also play a key role in many state of the art methods \citep{zhu2021neural,zhang2018link}. 

Our theoretical analysis relies on the assumption that initial node features are orthonormal which covers widely used initialization schemes such as as one-hot encodings and high dimensional random features, and also holds approximately for many empirical data sets with high dimensional features. Hence, our theoretical findings also provide new insights into the effectiveness of the widely used initialization schemes of one-hot encodings and high dimensional random features in graph representation learning. More generally our results show that untrained versions of message passing layers are highly amenable to theoretical analysis and hence could potentially serve as a general ansatz for the theoretical analysis of GNNs including settings beyond link prediction.

The main contributions of the paper are as follows:
\begin{enumerate}
    \item We show that untrained versions of widely used MP layers often outperform their fully trained counterparts in LP tasks. 
    \item We establish a direct connection between MPNNs and path based node similarity measures both of which are widely used in LP methods. 
    \item Our theoretical analysis further provides insights in to the effectiveness of widely used node initialization schemes such as one-hot-encodings and random features in graph neural networks.   
\end{enumerate}


\section{Related Work} \label{section: Related Work and Motivation}

Our work is motivated by recent works that investigate simplified and untrained GNNs from different perspectives. We formulate UTMP layers following the approach of \cite{wu2019simplifying}  which simplifies GNNs by successively removing trainable parameters from layers and nonlinearities between consecutive layers. \cite{wu2019simplifying} also provide a theoretical analysis of simplified models in the context of node classification reducing the model to a fixed low-pass filter followed by a linear classifier. The paper also empirically evaluates the simplified architectures on various downstream applications and shows that simplified architectures do not negatively impact accuracy while being computationally more efficient than their fully trained counterparts. 

Other works have focused on finding untrained subnetworks. For instance \cite{huang2022you} explores the existence of untrained subnetworks in GNNs that can match the performance of fully trained dense networks at initialization, without any optimization of the weights. The paper leverages sparsity as the core tool to find such subnetworks and shows that they can substantially mitigate the over-smoothing problem, hence enabling deeper GNNs. The paper also shows that the sparse untrained subnetworks have appealing performance in out-of-distribution detection and robustness to input perturbations. Similarly, in \cite{boker2023fine} the authors demonstrate that GNNs with randomly initialized weights, without training, can achieve competitive performance compared to their trained counterparts focusing on the problem of graph classification. In \cite{dong2023pure} the authors show that certain common neighbour measures can be approximated by MPNNs initialised with random weights and node features without training. Other more recent works on untrained GNNs include \cite{dong2024you} where the authors propose a training free linear GNN model for semi supervised node classification in text attributed graphs and  \cite{sato2024training} that defines training free GNNs for transductive node classification based on using training labels as features. 

Link prediction is widely studied problem with a multitude of available methods. UTMP layers are related to a both GNN based methods such as Variational Graph Autoencoders (V-GAE) \citep{kipf2016variational} and more traditional methods that are rely on path and random walk based measures for link prediction \citep{kumar2020link}. On the other hand state of the art methods such as SEAL \citep{zhang2018link}, NBFnet \citep{zhu2021neural}, BUDDY \citep{chamberlain2022graph}, Neo-GNN \citep{yun2021neo} and NCNC \citep{wang2023neural} in general rely on combining GNNs and structural features. Some methods such as SEAL \citep{zhang2018link} and WalkPool \citep{pan2021neural} are based on extracting and performing MP on local subgraphs around target links effectively framing link prediction as a graph classification problem. Although subgraph extraction based methods can out perform purely GNN based methods in link prediction tasks the subgraph extraction process can be resource intensive for large networks negatively affecting the scalability of these methods. NBFnet \citep{zhu2021neural} is another state-of-the-art method that is motivated by the Bellman-Ford algorithm. NBFnet is based on learning representations of paths between target nodes and aggregation functions for these representations. While NBFnet scales more favorably compared to subgraph extraction based methods it still needs to compute representations for large numbers of paths to predict links and hence has worse scaling behavior compared to purely GNN based link prediction methods \citep{zhu2021neural}. Neo-GNN \citep{yun2021neo} and BUDDY \citep{chamberlain2022graph} circumvent these difficulties by using pairwise similarity measures between higher order neighbourhoods of nodes, which are then used together with GNN based node features for LP. Notably BUDDY includes a feature propagation step that can be seen as a special case of  UTMP (see UTSAGE in Section \ref{section: Untrained MPNN Architectures}). In \cite{wang2023neural} the authors propose a GNN based approach that in addition to using node level features also aggregates features of common neighbours for link prediction and further propose Neural Common Neighbour Completion (NCNC) to counteract the negative effects of graph incompleteness on LP performance. 

As detailed above GNNs are widely employed as sub-components in LP methods. Hence rather than proposing a new method for link prediction we consider the advantages of using untrained MP layers over their trained counterparts in existing LP methods. Moreover, our theoretical results establish a link between MP based approaches and structural features by showing that features resulting from UTMP implicitly encode neighbourhood information that underlies many widely used common neighbour and path based structural features used in LP.

From a theoretical perspective, our results also relate to recent works on the effectiveness of random node initializations and one-hot encodings. For instance \cite{sato2021random} and \cite{abboud2020surprising} focus on the effect of random node initializations of the expressivity of GNNs in the context of graph classification while \cite{cui2022positional} explores various encodings for the task including node and graph classification. Our analysis complements these works from the point of view of LP and establishes a link between features derived from random and one hot initializations and path based topological features. In addition, complementary work on initialization strategies such as the graph partition based GPA initializer \cite{lin2020initialization} investigates how better initialization can improve downstream link prediction and node classification without changing the model architecture.

Finally, in our theoretical analysis we rely on the fact that collections of high dimensional vectors tend to be mutually orthogonal. This is a widely known fact that has wide-ranging applications in ML more broadly given the pervasive use of high dimensional vector representations in modern ML methods \citep{kanerva2009hyperdimensional,kanerva2018computing}. 

\section{Message Passing Architectures} \label{section: mpnn architecture}
Prior to introducing the message passing architectures investigated in our work, we first clarify the notation used throughout the paper.
Let $G(V,E)$ be an undirected graph with vertex set $V = \{v_1,v_2\ldots v_N\}$, edges $E \subseteq V \times V$ and no self-loops, i.e. $(v,v) \notin E \quad \forall v \in V$.
We denote the adjacency matrix of the graph as $A$ and define $\tilde{\mathbf{A}}:=\mathbf{A}+\mathbf{I}$, where $\mathbf{I}$ is the identity matrix, i.e. $\tilde{\mathbf{A}}$ denotes the adjacency matrix of the graph $G$ that explicitly includes all possible self-loops.
We use $\mathcal{N}(v)$ to denote the neighborhood of a node $v$, i.e. the set $\{ w \in V: (v,w) \in E\}$, and use $\tilde{\mathcal{N}}(v):=\mathcal{N}(v) \cup \{v\} $ to denote the neighborhood of $v$ in the graph with self-loops.
Similarly, we denote the degree of a node $v$ as $d(v)$ and $\tilde{d}(v)=d(v)+1$. 
The initial feature vector of node $v$ is denoted as $h_v^{(0)}$ and we use $h_v^{(l)}$ to denote the updated feature vector of node $v$ after $l$ rounds of message passing. Although we restrict our discussion to undirected and unweighted graphs the generalization of our definitions and results to weighted graphs is straightforward.

Prior to defining untrained versions, we first introduce the message passing rules of the four GNN architectures considered in our work, using a unified notation above.

\paragraph{On the inclusion of self-loops} Throughout our formulations we explicitly include self-loops, i.e., work with $\tilde{\mathcal{N}}(v)=\mathcal{N}(v)\cup\{v\}$ and $\tilde{\mathbf{A}}=\mathbf{A}+\mathbf{I}$. This choice is motivated by both theoretical and practical considerations. First, self-loops allow nodes to retain their own features during aggregation, which prevents a complete dilution of node-specific information as the number of layers increases. Second, self-loops enable UTMP layers to capture paths of all lengths up to $2l$ (rather than exactly length $2l$), which makes the connection to path-based measures such as Katz index and rooted PageRank more direct. Third, this aligns our untrained formulations with standard trained architectures (e.g., GCN, SAGE, GraphConv), ensuring a fair comparison when removing trainable parameters.

\paragraph{Graph Convolutional Networks \citep{kipf2016semi}}
GCNs were introduced as a scalable approach for semi-supervised learning on graph-structured data. GCNs are based on an efficient variant of convolutional neural networks which operate directly on graphs. The MP layer of GCN is given by:
$$
h_v^{(l)} = W^{(l)}\cdot \sum_{u \in \tilde{\mathcal{N}}(v)} \frac{1}{\sqrt{\tilde{d}_{u}\tilde{d}_{v}}}h_u^{(l-1)},
$$
where $W^{(l)}$ is the weight matrix for layer $l$. 

\paragraph{GraphSAGE \citep{hamilton2017inductive}}
GraphSAGE is a type of Graph Neural Network that uses different types of aggregators such as mean, gcn, pool, and lstm to aggregate information from neighboring nodes. 
The MP layer in GraphSAGE uses the following formula:
$$
h_v^{(l)} = W^{(l)}_1\cdot h_v^{(l-1)} + W^{(l)}_2 \cdot \text{AGG}_{u \in \mathcal{N}(v)} h_u^{(l-1)},
$$
where \(W^{(l)}_{\{1,2\}}\) are learned weight matrices at layer \(l\), \(AGG\) is an aggregation function (such as mean, sum, max).

Throughout this paper use the following slightly modified version of the SAGE layer :
$h_v^{(l)} = W_1^{(l)} \cdot \frac{1}{\tilde{d}_v}\sum_{u \in \tilde{\mathcal{N}}(v)} h_u^{(l-1)},$
which we found to produce superior results for link prediction.

\paragraph{GIN \citep{xu2018powerful}}
The Graph Isomorphism Network Convolution (GIN) is a simple architecture that is provably the most expressive among the class of GNNs and is as powerful as the Weisfeiler-Lehman graph isomorphism test. The MP step of GIN is as follows:
$$
h_v^{(l)} = \Theta\big( (1 + \epsilon) \cdot h_v^{(l-1)} + \sum_{u \in \mathcal{N}(v)} h_u^{(l-1)}  \big),
$$
where \(\Theta\) denotes an MLP after each message passing layer, which in our implementation includes two Linear layers and a Rectified Linear Unit (ReLU) activation function following each Linear layer (code adapted from \cite{boker2023fine, finegranedmorris}).

\paragraph{GraphConv \citep{morris2019weisfeiler}}
GraphConv is a generalization of GNNs, which can take higher-order graph structures at multiple scales into account. The mathematical formulation of this is as follows:
$$
h_v^{(l)} = W_1^{(l)} \cdot h_v^{(l-1)} + W_2^{(l)} \cdot\sum_{u \in \mathcal{N}(v)}  h_u^{(l-1)}
$$ where \(W^{(l)}_{\{1,2\}}\) are learned weight matrices.

\subsection{Untrained MP Architectures} \label{section: Untrained MPNN Architectures}

For the purpose of our theoretical and experimental evaluation, we now define the untrained counterparts of the four Message Passing Neural Network (MPNN) architectures introduced in the previous section.
Following \cite{wu2019simplifying} we eliminate all learnable components and replace them with identity matrices.
Here our objective is to obtain the simplest form for the update function that retains the general message passing strategy, which includes the predefined update message passing functions and aggregation methods while removing all learnable parameters and nonlinearities.
We obtain the following functions that capture the aggregation and update step in the untrained versions of the message passing layers:
\begin{enumerate}
    \item[] \textbf{UTGCN:} $h_v^{(l)} = \sum_{u \in \tilde{\mathcal{N}}(v)} \frac{1}{\sqrt{\tilde{d}_{u}\tilde{d}_{v}}}h_u^{(l-1)}$
    \item[] \textbf{UTSAGE:} $h_v^{(l)} = \frac{1}{\tilde{d}_v}  \sum_{u \in \tilde{\mathcal{N}}(v) } h_u^{(l-1)}$
    \item[] \textbf{UTGIN:} $h_v^{(l)} =  (1 + \epsilon)  h_v^{(l-1)} + \sum_{u \in \mathcal{N}(v)} h_u^{(l-1)}$
    \item[] \textbf{UTGraphConv:} $h_v^{(l)} =  \sum_{u \in \tilde{\mathcal{N}}(v)}  h_u^{(l-1)}$
\end{enumerate}

In general, we consider the case where all nodes have self-loops, i.e. the features of the node itself are included in the aggregation step.
Further setting $\epsilon = 0$ for GINs results in a uniform formula across both models: $h_v^{(l)} =   \sum_{u \in \tilde{\mathcal{N}}(v)} h_u^{(l-1)}$.
Henceforth we will refer to both models as UTGIN.

The simplified message passing layers can also be expressed in matrix form: $$\mathbf{H}^{(l)}=\mathbf{S}\mathbf{H}^{(l-1)}=\mathbf{S}^l\mathbf{H}^{(0)}, $$
where $\mathbf{H}^{(0)} \in \mathbb{R}^{n\times d}$ is the initial feature matrix, and $\mathbf{H}^{(l)}$ the feature matrix after $l$ iterations of message passing. Following, the definitions of UTMP layers above we have $\mathbf{S}=\tilde{\mathbf{D}}^{-1/2}\tilde{\mathbf{A}}\tilde{\mathbf{D}}^{-1/2}$ for UTGCN and  $\mathbf{S}=\tilde{\mathbf{D}}^{-1}\tilde{\mathbf{A}}$ for UTSAGE, where $\tilde{\mathbf{D}}$ is the degree matrix with diagonal entries $\tilde{\mathbf{D}}_{uu}=\sum_v \tilde{\mathbf{A}}_{uv}$. Similarly, for UTGIN we have $\mathbf{S}=\tilde{\mathbf{A}}$. The generalization of UTMP layers to undirected weighted graphs can be obtained by simply replacing the adjacency matrix and related quantities with their weighted counterparts in the formulation of $\mathbf{S}$. 

\subsection{Simplified architectures} 
\label{section: Simplified MPNNs}
Following the construction of \cite{wu2019simplifying} for the case of node classification we add a final trained linear layer before the final dot product. We refer to such architectures that include a final trained linear layer after the UTMP layers as 'simplified' in accordance with \cite{wu2019simplifying} and include an 'S' in the abbreviations of these models, e.g. SGCN. This results in an architecture where the final node features  are given by: 
$$\mathbf{H}^{(l)}=\Theta\mathbf{S}^l\mathbf{H}^{(0)},$$
where $\Theta$ is the learned weight matrix of the linear layer. In the case of simplified GNN architectures, the trained linear layer can also be interpreted as a modified positive semi-definite inner product in the form of $\langle \Theta h_v^l, \Theta h_u^l\rangle$ where $\Theta$ is the weight matrix of the linear layer. 

In the case of link prediction features produced by UTMP layers can actually be used to construct fully untrained architectures that only consist of feature propagation steps followed by an inner product. In practice, we found that such architectures based solely on UTMP layers can do surprisingly well in terms of LP performance showing that UTMP layers produce highly informative features. 

\subsection{UTMP layers and path based measures}
\label{section:multi MP}

Building on the formulations of the untrained layers above, in the following we provide a theoretical analysis that relates the inner products of features resulting from untrained message passing layers to pair-wise measures of node similarity that are based on characteristics of \emph{paths} in the underlying graph. Such path based measures offer a way of quantifying the indirect connection strength between node pairs in the absence of a direct link connecting the nodes. In order to relate path based measures and UTMP layers we will assume that initial feature vectors are pairwise orthonormal i.e. $\langle h_v^{(0)}, h_u^{(0)}\rangle = \delta_{u,v}$. 

A path of length $l$ is  defined as a sequence of $l+1$ vertices $(v_0,v_1 \ldots v_l)$ such that $(v_i,v_{i+1})\in E$ for all $0 \leq i < l$. We denote the space of a set of all paths of length $l$ between nodes $u$ and $v$ as $P^{l}_{uv}$. The number of paths of length $l$ between any $u$ and $v$ is given by the $l^{th}$ power of the adjacency matrix i.e. $|P^{l}_{uv}|=\tilde{\mathbf{A}}^l_{uv}$. Note that since we assume self-loops on all vertices $P^{l}_{uv}$ implicitly also includes shorter paths between $u$ and $v$. Similarly, paths of length $l$ between vertices $u$ and $v$ also determine the probability of a random walk starting at $u$ reaching $v$ which is given by $P(u\xrightarrow{l} v)= \sum_{p\in P^{l}_{uv}} \prod_{i \in p-[v]} \frac{1}{\tilde{d}_i}$, where $p-[v]$ denotes that the last vertex ($v$) is not included in the product. The random walk probability can also be expressed in matrix form  $P(u\xrightarrow{l} v)=(\tilde{\mathbf{D}}^{-1}\tilde{\mathbf{A}})^l_{uv}$. 

Now we consider inner products of features after $l$ iterations of message passing which is given by $\langle h_u^{(l)},h_v^{(l)}\rangle=(\mathbf{S}^l\mathbf{H}^{(0)}\mathbf{H}^{(0)\top}(\mathbf{S}^l)^\top)_{uv}$. For 
orthonormal features the inner products of features reduces to $\mathbf{H}^{(0)}\mathbf{H}^{(0)\top}=\mathbf{I}$ and we obtain the following expression for the inner product of the features after $l$ iterations of UTMP layers:
$$\langle h_u^{(l)},h_v^{(l)}\rangle=(\mathbf{S}^l(\mathbf{S}^l)^\top)_{uv}.$$

For UTGCN we have $\mathbf{S}=\tilde{\mathbf{D}}^{-1/2}\tilde{\mathbf{A}}\tilde{\mathbf{D}}^{-1/2}$ and the inner product after $l$ layers can be expressed in terms of paths of length $2l$ between $u$ and $v$ as: 
$$\langle h_u^{(l)},h_v^{(l)}\rangle =\frac{1}{\sqrt{\tilde{d}(u) \tilde{d}(v)}}\sum_{p\in P^{2l}_{uv}} \prod_{i \in [p]} \frac{1}{\tilde{d}_i},$$
where $[p]$ denotes the path $p$ with the first and last vertices removed. The above expression is equivalent to $\sqrt{P(u\xrightarrow{2l} v)P(v\xrightarrow{2l} u)}$ i.e. the geometric mean of the probabilities that a random walk starting at either $u$ or $v$ reaches the other in $2l$ steps. Similarly, for UTSAGE we have $\mathbf{S}=\tilde{\mathbf{D}}^{-1}\tilde{\mathbf{A}}$ and the inner product can be expressed in terms of paths in $P^{2l}_{uv}$  as: 
$$\langle h_u^{(l)},h_v^{(l)}\rangle =\sum_{p\in P^{2l}_{uv}} \prod_{i \in p -m(p)}  \frac{1}{\tilde{d}_i},$$
where $m(p)$ is the midpoint of the path $p$. The above expression is equivalent to $\langle h_u^{(l)},h_v^{(l)}\rangle = \sum_i P(u\xrightarrow{l} i)P(v\xrightarrow{l} i)$ and hence corresponds to the probability that two simultaneous random walks starting at $u$ and $v$, respectively, meet after $l$ steps at some midpoint. Finally, for UTGIN we have $\mathbf{S}=\tilde{\mathbf{A}}$ and hence: 
$$\langle h_u^{(l)},h_v^{(l)}\rangle =|P^{2l}_{uv}|.$$

Although the condition of orthonormality might seem quite restrictive at first glance it applies in many practical settings, though in some cases only approximately. Moreover, for the above result to hold orthogonality only needs to be satisfied in the common $l$-neighbourhood of the nodes. One example of orthonormal features that are widely used in practice are one hot encodings and orthonormality also applies in the case of high dimensional random feature vectors since for sufficiently large dimensions any set of $k$ independent random vectors is quasi orthogonal \citep{eaton2007random}. Similar results also hold for random high dimensional binary features that are sparsely populated for which the expected value of the inner product of two vectors scales as $O(1/k)$ for dimension $k$.

High dimensional features of empirical data sets also show similar characteristics to their random counterparts. For instance, empirical feature vectors of randomly selected node pairs tend to be approximately orthogonal, notwithstanding the fact that features of connected node pairs can be highly correlated \citep{nt2019revisiting}, as can be verified experimentally (see Sec.\ref{section: feature dot products}).

As mentioned before in the case of simplified GNN architectures, the final trained linear layer can be interpreted as a modified positive semi-definite inner product and the orthogonality results for high dimensional random features also apply to such more general inner products. However, note that normality is no longer guaranteed i.e. $\langle  \Theta h_v, \Theta h_u\rangle  \sim \delta_{u,v} |h_u|^2_{\Theta}$.

We would like to note that the assumption of orthogonality is a mathematical assumption we use to establish the connection between UTMP layers and path based measures. However, this does not imply that UTMP require orthogonal features to perform well at LP tasks. On the contrary deviations from orthogonality can enhance the LP performance of UTMP for instance when connected nodes tend to have more similar features which holds for many LP benchmarks.  We find that the inner products of feature vectors of randomly selected node pairs are in general close to zero. Note that, the feature vectors of all datasets are non-negative as they represent word occurrences. As expected, for connected nodes the inner products of feature vectors tend to be higher reflecting the increased feature similarity. 

\begin{figure}[!htp]
    \centering
    \includegraphics[width=0.7\textwidth]{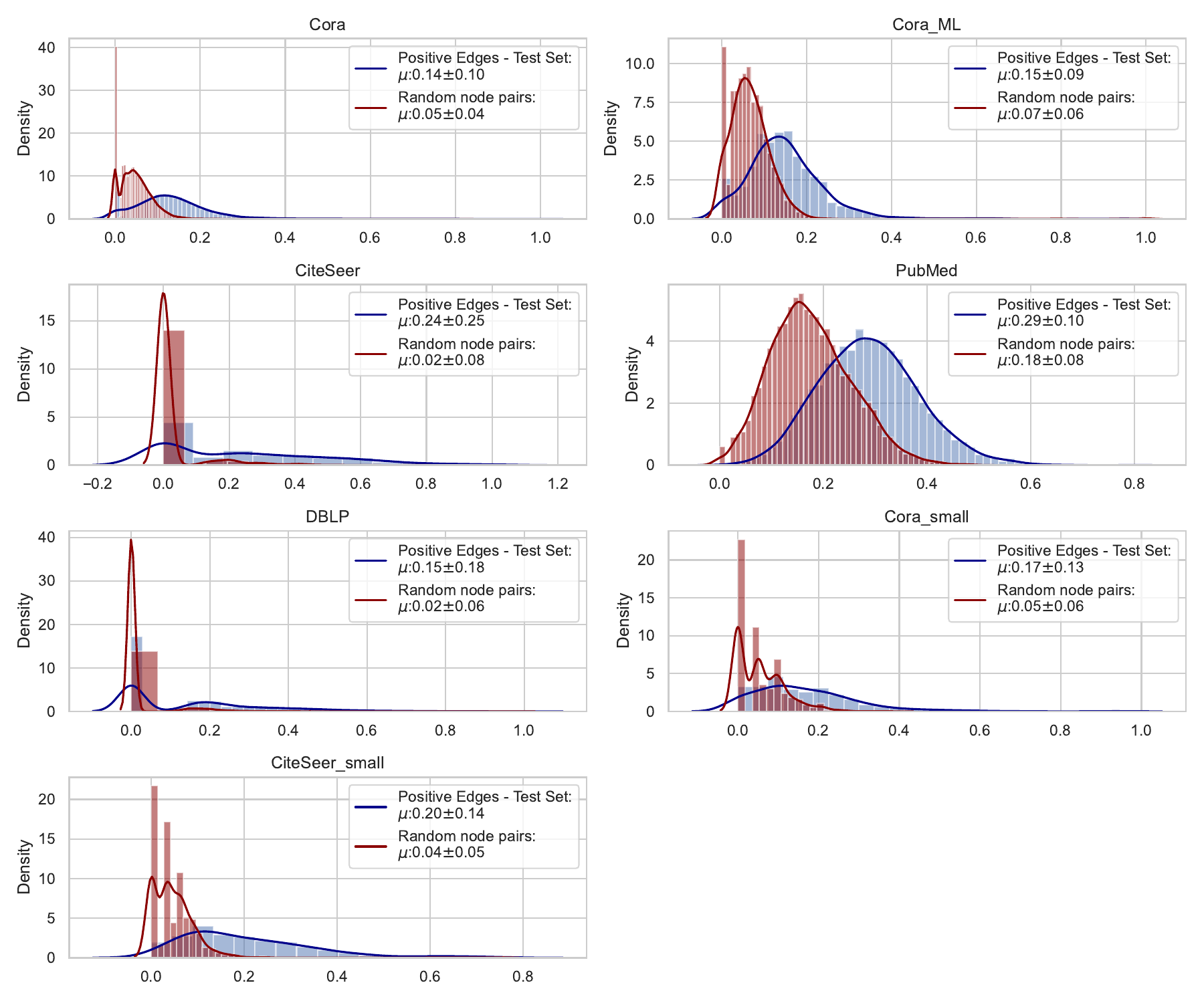}
    \caption{The distribution of feature dot products for pairs of connected and random node pairs for the attributed datasets.}
    \label{fig:dot product}
\end{figure}

When initial features deviate from exact orthogonality, we can write their Gram matrix as $\mathbf{H}^{(0)}\mathbf{H}^{(0)\top}=\mathbf{I}+\Delta$, where $\Delta$ captures correlations in the initial features. In this case, the UTMP similarity decomposes as
\begin{equation*}
\mathbf{S}^l\mathbf{H}^{(0)}\mathbf{H}^{(0)\top}(\mathbf{S}^l)^{\top}=\underbrace{\mathbf{S}^l(\mathbf{S}^l)^{\top}}_{\text{path-based term}}+\underbrace{\mathbf{S}^l\Delta(\mathbf{S}^l)^{\top}}_{\text{feature similarity term}}\,.
\end{equation*}
The first term is the pure path-based quantity; the second term reflects the effect of feature similarity transported by $\mathbf{S}^l$. Under homophily, this additional term is often beneficial for LP; conversely, large misaligned correlations (large $\Delta$) can introduce noise. Empirically, we find inner products between random node pairs concentrate near zero (Fig.\,\ref{fig:dot product}), suggesting that $\Delta$ is typically small in our datasets.

\subsection{Triadic closure and other path based measures}
\label{section: Triadic closure and other path based measures}

Triadic closure, also known as transitivity, refers to the tendency for nodes in real-world networks to form connections if they share (many) common neighbors. As such triadic closure has been widely studied as a mechanism that drives link formation in complex real-world networks \citep{rapoport1953spread,holland1971transitivity}.
Moreover, node similarity measures that build on triadic closure in social networks have been used for similarity-based link prediction algorithms \citep{LU20111150}.

Given a pair of nodes $(u,v)$, the tendency of them to be connected due to triadic closure can be quantified by simply counting the number of common neighbours between the two vertices i.e. $T(u,v)=|N(u) \cap N(v)|$ which corresponds to $l=1$ for UTGIN, assuming that $u$ and $v$ are not connected in the graph as is typically the case in an LP setting. In practice, one might further want to account for the fact that in general nodes with higher degrees also have a larger probability of having common neighbours, for instance by normalizing by the degrees, i.e.:  $T_d(u,v)=|N(u) \cap N(v)|/\tilde{d}(u)\tilde{d}(v)$, which corresponds to $l=1$ for UTSAGE. One can go one step further and also take into account the degrees of the common neighbours themselves since high degree nodes are by definition common neighbours of more node pairs, for instance by weighing common neighbours according to their degree $T_n(u,v)=\frac{1}{\sqrt{\tilde{d}(u) \tilde{d}(v)}}\sum_{i \in N(u) \cap N(v)} \frac{1}{\tilde{d}_i}$ which in our case corresponds to UTGCN with $l=1$. 

Our results also link UTMP layers to other topological similarity measures that are widely used in link prediction heuristics such as the Adamic-Adar (AA) index \citep{adamic2003friends}, Resource Allocation (RA) \citep{zhou2009predicting}, the Katz index \citep{katz1953new}, rooted PageRank \citep{brin1998anatomy} and SimRank \citep{jeh2002simrank}. For instance the AA index,  given by $AA(u,v)=\sum_{i \in N(u) \cap N(v)} \frac{1}{\log \tilde{d}_i }$, and $RA(u,v)=\sum_{i \in N(u) \cap N(v)} \frac{1}{\tilde{d}_i}$ differ only slightly from the triadic closure measures we obtained for UTMP layers. Similar results also hold for other path based measures such as rooted PageRank, the Katz index and SimRank which can be defined in terms of power series over paths of different lengths. For instance, SimRank similarity between nodes $u$ and $v$ is defined as $s(x,y)=\sum_l P_{uv}(l)\gamma^l$ where $P_{uv}(l)$ is the probability that two random walks starting at $u$ and $v$ meet after $l$ steps and $0<\gamma<1$ is a free parameter. Similarly the Katz index is defined as $Katz(u,v)=\sum_l \mathbf{A}^l_{uv}\gamma^l$ and rooted PageRank is defined as $PR(u,v)=(1-\gamma)\sum_l \frac{P(u\xrightarrow{l} v)+P(v\xrightarrow{l} u)}{2} \gamma^l$ again with $0<\gamma<1$ being a free parameter. Hence, the Katz index is closely related to UTGIN and rooted PageRank is closely related to UTGCN, the main difference being that these measures also include paths of odd length which UTGIN and UTGCN include only indirectly through the inclusion of self loops in their formulation.

\section{Experiments and Results} \label{section: experiments and results}

In the following, we provide details on our experimental setup.
We evaluate GNN architectures on a variety data sets that cover both attributed graphs where nodes have additional high-dimensional features (Cora small, CiteSeer small, Cora, CoraML, PubMed, CiteSeer, DBLP) and non-attributed graphs that do not contain any node features. Data sources and summary statistics of the data sets can be found in the Appendix Table \ref{table: dataset description}. We use the area under the Receiver Operator Characteristic curve (ROC-AUC) for the non-attributed datasets and Hits@100 and ROC-AUC for the attributed datasets as our main performance measures.\footnote{The code for replicating the results is available at: \url{https://doi.org/10.5281/zenodo.15019863}}

\subsection{Experimental Setup} \label{section: Experimental Setup}

To ensure a fair comparison among models we maintain the same overall architectures across all experiments and MP layers. For trainable message passing layers, each layer is followed by an Exponential Linear Unit (ELU) and the optimal number of layers for models is determined via hyperparameter search. Upon completion of the message passing layers, we introduce a final linear layer for both trained and simplified models. We also consider untrained (UT) models that do not include this final linear layer and directly take the inner product between the propagated features of the source and target nodes resulting in a parameter-free and hence fully untrained model. Since the simplified architectures consist of UTMP layers followed by a trainable linear layer, the consideration of UT models which do not include the linear layer also covers all possible ablation studies. 

In principle any LP method that uses GNNs as one of its sub-components can also be formulated using UTMP layers. However, in general state of the art methods consist of many sub-components resulting in more complex and computationally demanding experimental setups where the effect of switching from trained to untrained MP layers is difficult to isolate. Therefore, we focus mostly on graph autoencoders in our experiments due to their simplicity, but also consider two versions of NCNC \cite{wang2023neural}: one which uses trained GNN layers for MP and another that uses UTGNN layers instead, which following our naming convention is denoted as SNCNC. 

For the simplified models we precompute node features corresponding to the untrained message passing layers, as these do not change during training. 
We use one-hot encoding as initial node features for the non-attributed datasets. Further details about the experimental setup can be found in the supplementar material (Sec. \ref{section: Hyperparameter choices}) along with results based on HeaRT \cite{li2023evaluating} evaluation setting (Sec. \ref{sec; Heart split}) which samples hard negative samples via multiple heuristics.

\begin{table}[!htbp]
\centering 
\caption{Link Prediction accuracy for attributed networks as measured by Hits@100. Red values correspond to the overall best model for each dataset, and blue values indicate the best-performing model within the same category of message passing layers.}
\label{table: results}
\resizebox{\textwidth}{!}{%
\begin{tabular}{lccccccc}
\textbf{Models}& \textbf{Cora (small)} & \textbf{CiteSeer (small) } & \textbf{Cora } & \textbf{Cora ML} & \textbf{PubMed} & \textbf{CiteSeer}    & \textbf{DBLP}  \\ \hline  \hline&   \textit{Hits@100} & \textit{Hits@100} & \textit{Hits@100} & \textit{Hits@100} & \textit{Hits@100} & \textit{Hits@100}    & \textit{Hits@100} \\ \hline  \hline
GCN &80.5 ± 1.59&83.0 ± 1.57&79.75 ± 0.74&82.92 ± 1.44&\textcolor{blue}{73.64 ± 2.04}&81.06 ± 1.3&\textcolor{blue}{69.72 ± 1.71}   \\
SGCN &\textcolor{blue}{84.73 ± 1.46}&\textcolor{blue}{88.69 ± 0.57}&\textcolor{blue}{83.93 ± 0.8}&\textcolor{blue}{87.31 ± 1.23}&69.11 ± 1.25&\textcolor{blue}{86.02 ± 1.11}&64.81 ± 2.09    \\ 
UTGCN &64.24 ± 3.4&81.07 ± 1.5&37.5 ± 1.5&58.1 ± 1.74&25.35 ± 2.04&69.39 ± 2.22&31.46 ± 1.04  \\ \hline
SAGE &75.67 ± 1.29&80.23 ± 1.09&69.07 ± 1.05&78.33 ± 0.85&\textcolor{blue}{56.25 ± 0.7}&77.14 ± 2.93&\textcolor{blue}{64.81 ± 1.66}   \\
SSAGE &\textcolor{blue}{80.42 ± 1.71}&\textcolor{blue}{87.22 ± 1.19}&\textcolor{blue}{74.16 ± 1.47}&\textcolor{blue}{79.61 ± 1.75}&42.14 ± 1.85&\textcolor{blue}{83.78 ± 1.61}&56.49 ± 2.64   \\  
UTSAGE &57.76 ± 1.51&61.85 ± 3.23&30.51 ± 1.57&51.13 ± 1.27&6.6 ± 0.75&69.88 ± 2.15&19.04 ± 2.2    \\ \hline
GIN &\textcolor{blue}{74.66 ± 1.63}&71.16 ± 1.67&69.83 ± 1.07&78.61 ± 1.07&\textcolor{blue}{65.3 ± 1.3}&74.64 ± 1.61&64.81 ± 1.66   \\ 
GraphConv &74.7 ± 1.14&74.89 ± 1.59&62.37 ± 1.87&\textcolor{blue}{78.66 ± 1.57}&62.84 ± 2.1&77.69 ± 1.32&66.59 ± 1.25     \\ \hdashline
SGIN &74.54 ± 1.69&\textcolor{blue}{78.71 ± 2.15}&\textcolor{blue}{73.11 ± 1.03}&77.46 ± 1.77&46.21 ± 0.85&\textcolor{blue}{78.56 ± 1.31}&\textcolor{blue}{66.59 ± 1.15} \\
UTGIN &46.73 ± 2.36&61.85 ± 3.23&22.65 ± 1.13&44.79 ± 1.44&22.01 ± 1.71&58.8 ± 6.18&34.29 ± 1.02  \\ \hline \hline
NCNC & 83.69±3.13 & 76.37±2.90 & 84.55±1.14 & 87.36±1.83 & 80.72±0.91 & 87.22±3.61 & 73.77±0.75 \\ 
SNCNC & \textcolor{red}{88.72±1.20} & \textcolor{red}{93.42±0.78} & \textcolor{red}{84.69±1.39} & \textcolor{red}{89.81±0.86} & \textcolor{red}{81.26±1.59} & \textcolor{red}{89.79±1.51} & \textcolor{red}{74.23±0.117}  

   \\ 
   \hline \hline

\end{tabular}%
}
\end{table}

\subsection{Experimental Results} \label{section: Experimental Results}

In the following section, we discuss the results of our experiments for link prediction in graphs with node attributes (i.e. in graphs where nodes have additional features) and non-attributed graphs separately.
This diverse selection of data sets allows us to thoroughly evaluate the capabilities of the models for graphs from different application scenarios, with different sizes, and different topological characteristics.

Results for attributed graphs are given in Table \ref{table: results} where we find that in general replacing trained MP layers with their untrained counterparts increases LP performance  on most datasets with the exception of PubMed and DBLP data sets where architectures based on UTMP layers perform worse in terms of Hits@100. In general, we find that GCN based architectures have the best overall performance. Finally, we observe that replacing trained GCN layers with their untrained counterpart also improves the LP performance of NCNC. Indeed, some of the results reported in \cite{wang2023neural} seem to be obtained using UTGCN layers.
Although we focus is on MP layers rather end-to-end LP methods, we find that simple GAE type architectures based on UTMP layers can in many cases outperform more sophisticated state-of-the-art models such as SEAL, NBFnet and Neo-GNN (Sec. \ref{sec; Heart split}).

We also find that the fully untrained (UT) architectures already provide a very good baseline and some cases even outperform fully trained versions in terms of ROC-AUC (Table \ref{table: results roc}). This demonstrates that the raw features produced by UTMP layers, which the simplified models are trained on, are already highly informative for link prediction in accordance with our theoretical results. Note that, the fully untrained (UT) models can be computed efficiently via sparse matrix multiplication. 

In our analysis of the OGB datasets, we found that NCNC (GIN) outperforms other models in two out of three datasets. In the remaining dataset, SNCNC (SGIN) showed superior performance compared to the other models. Additionally, we observed that SNCNC models are highly competitive with fully trained models and, in some cases, are less memory-intensive. For instance, while the NCNC (GCN) model ran out of memory, the SNCNC (SGCN) model produced very good results without encountering this issue.

In the case of non-attributed graphs (Table \ref{table: AUC results noattr}) we observe that models based on UTMP layers achieve the highest score on 6 out of 8 datasets, with the exceptions being NS and Router datasets.  Moreover, we find that the fully untrained UTGCN model performs best on the 'Celegans', 'PB', 'USAir', 'E-coli' which can be attributed to the reduced dimension of the learned features that come with the linear layers present in the simplified and fully trained models. Furthermore, as we used one hot encodings as initial node features for the unattributed datasets orthonormality is satisfied exactly and therefore there is a one-to-one correspondence between the UT models and path based topological measures. 

\begin{table}[!htbp]
\centering 
\caption{Link Prediction for non-attributed networks as measured by ROC-AUC. }
\label{table: AUC results noattr}
\resizebox{\textwidth}{!}{%
\begin{tabular}{lcccccccc}
\textbf{Models}& \textbf{NS} & \textbf{Celegans} & \textbf{PB} & \textbf{Power}    & \textbf{Router}  & \textbf{USAir} & \textbf{Yeast} & \textbf{E-coli}\\ \hline  \hline
GCN &\textcolor{blue}{95.22 ± 1.8}&87.98 ± 1.45&92.91 ± 0.3&74.68 ± 2.67&\textcolor{red}{91.42 ± 0.44}&93.56 ± 1.53&94.49 ± 0.61&98.48 ± 0.22 \\
SGCN &95.17 ± 0.96&89.38 ± 1.42&93.86 ± 0.42&\textcolor{blue}{81.08 ± 1.2}&77.51 ± 1.85&94.08 ± 1.43&\textcolor{red}{95.74 ± 0.33}&98.32 ± 0.2 \\ 
UTGCN &94.76 ± 1.03&\textcolor{red}{91.47 ± 1.4}&\textcolor{blue}{94.49 ± 0.38}&72.97 ± 1.27&61.68 ± 1.01&\textcolor{red}{94.81 ± 1.1}&94.0 ± 0.43&\textcolor{blue}{99.37 ± 0.1} \\ \hline
SAGE &\textcolor{red}{95.9 ± 0.86}&87.32 ± 1.61&\textcolor{blue}{92.94 ± 0.57}&74.17 ± 2.03&62.6 ± 3.3&\textcolor{blue}{93.37 ± 1.2}&94.43 ± 0.67&\textcolor{blue}{98.22 ± 0.13} \\
SSAGE &95.21 ± 1.09&\textcolor{blue}{88.05 ± 1.8}&91.66 ± 0.43&\textcolor{red}{81.84 ± 1.49}&\textcolor{blue}{70.1 ± 1.3}&92.25 ± 1.45&\textcolor{blue}{95.72 ± 0.31}&93.59 ± 0.14 \\
UTSAGE &94.72 ± 1.07&84.48 ± 1.87&86.46 ± 0.64&72.96 ± 1.26&61.47 ± 0.99&87.94 ± 1.58&93.45 ± 0.45&85.56 ± 0.37 \\ \hline
GIN &95.24 ± 1.22&86.74 ± 2.3&93.04 ± 0.99&71.97 ± 2.3&\textcolor{blue}{87.84 ± 3.05}&92.14 ± 0.98&94.7 ± 0.45&\textcolor{blue}{98.43 ± 0.24} \\
GraphConv &\textcolor{blue}{95.73 ± 1.4}&86.64 ± 2.31&92.99 ± 0.87&\textcolor{blue}{74.31 ± 1.93}&80.84 ± 1.28&91.16 ± 1.76&94.94 ± 0.38&98.32 ± 0.22 \\ \hdashline
SGIN &95.48 ± 0.88&\textcolor{blue}{88.31 ± 1.3}&\textcolor{blue}{93.72 ± 0.48}&73.73 ± 1.69&72.83 ± 1.28&93.02 ± 1.37&\textcolor{blue}{95.63 ± 0.49}&97.68 ± 0.2 \\
UTGIN &94.62 ± 1.05&86.48 ± 1.29&92.77 ± 0.51&72.93 ± 1.27&61.67 ± 1.02&\textcolor{blue}{93.44 ± 0.84}&92.94 ± 0.41&95.81 ± 0.22 \\ \hline 
NCNC &\textcolor{blue}{92.66  ± 1.94}& 86.01 ± 3.13  & 95.27 ± 0.26  & 61.63 ± 2.18 & 73.06 ± 2.96 & 91.10 ± 2.14 & 93.41 ± 0.46 & 99.53 ± 0.09 \\
SNCNC & 91.28 ± 2.97 & \textcolor{blue}{ 88.18 ± 2.65 } & \textcolor{red}{ 95.77 ± 0.22 } & \textcolor{blue}{ 68.41 ± 1.46  } & \textcolor{blue}{ 87.29 ± 1.20 } & \textcolor{blue}{ 93.95 ± 1.36} & \textcolor{blue}{95.42 ± 0.4} & \textcolor{red}{ 99.62 ± 0.05}
\\\hline  
\hline
\end{tabular}%
}

\end{table}

\begin{figure}[!htb]
    \centering
    \includegraphics[width=0.8\textwidth]{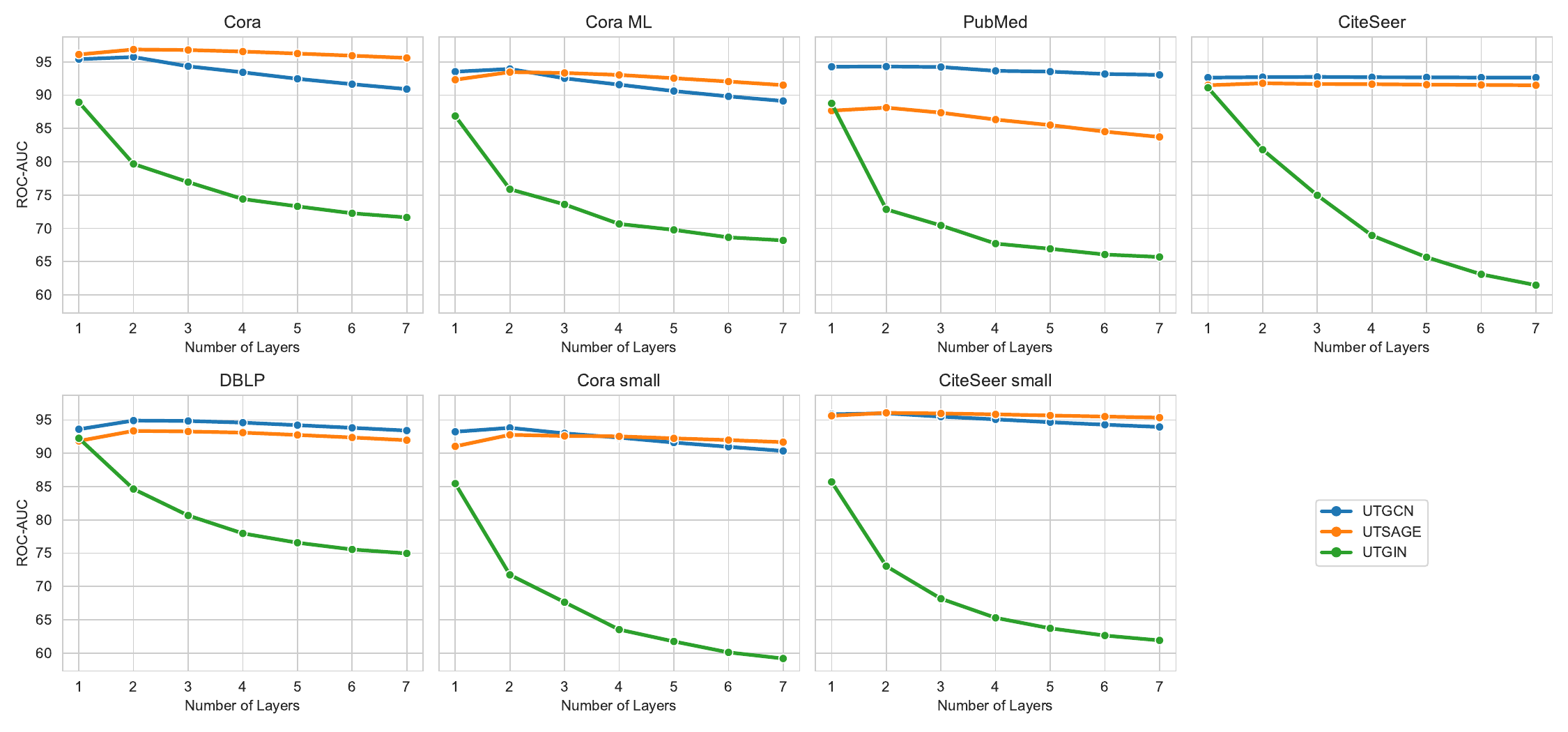}
    \caption{The effect of increased layer size for fully untrained models.}
    \label{fig:multi MP}
\end{figure}

Finally, we also examine the effect of increasing the number of UTMP layers using fully untrained (UT) models. Our results in Fig.\ref{fig:multi MP} indicate that, in general, UTGCN and UTSAGE maintain their performance as the number of layers is increased whereas the performance of UTGIN decreases sharply with more layers. This behavior can be attributed to the lack of degree based normalization in the formulation of GIN (see Sec.\ref{section:multi MP}) which leads UTGIN to be dominated by longer paths, and hence longer distance correlations, as the number of layers increases. In general however we find that UTMP layers do not suffer from over-squashing when equipped with proper degree based normalisation which can be attributed to the absence of nonlinearities and mixing between feature dimensions in UTMP layers. 

\section{Conclusion} \label{section: conclusion}

In this work, we explored the application of graph neural networks with untrained message passing layers for link prediction. 
Interestingly, our experimental evaluation shows that simplifying GNN architectures by eliminating trainable parameters and nonlinearities can enhance link prediction performance and training efficiency.
As such, untrained message passing layers offer a computationally efficient alternative to their fully trained counterparts that naturally scales to large graphs, while providing clearer theoretical links to classical path-based heuristics.
To complement our experimental results, we offered a theoretical perspective on untrained message passing, analytically establishing links between features generated by untrained message passing layers and path-based topological measures. 
We found that the link prediction offers a complementary perspective for analysing MPNNs and provides insights into the topological features captured by widely used initialization schemes such as random features and one-hot encodings. 

In future work, we hope to extend our study to other classes of graphs, such as directed, signed, weighted, and temporal networks. 
The conceptual simplicity of untrained message passing layers might also be a useful guide in designing new graph neural network architectures or adapting existing architectures to directed or temporal networks.
We thus believe that our work is of interest both for the community of researchers developing new machine learning methods, as well as for practitioners seeking to deploy efficient and resource-saving models in real-world scenarios.

\section*{Acknowledgements}
Lisi Qarkaxhija and Ingo Scholtes acknowledge funding from the German Federal Ministry of Education and Research (BMBF) via the Project "Software Campus 3.0", Grant No. (FKZ) 16IS24030. 

\bibliographystyle{unsrtnat}
\bibliography{log_2025}

\begin{thebibliography}{56}
\providecommand{\natexlab}[1]{#1}
\providecommand{\url}[1]{\texttt{#1}}
\expandafter\ifx\csname urlstyle\endcsname\relax
  \providecommand{\doi}[1]{doi: #1}\else
  \providecommand{\doi}{doi: \begingroup \urlstyle{rm}\Url}\fi

\bibitem[Basu et~al.(2022)Basu, Gallego-Posada, Vigan{\`o}, Rowbottom, and Cohen]{basu2022equivariant}
Sourya Basu, Jose Gallego-Posada, Francesco Vigan{\`o}, James Rowbottom, and Taco Cohen.
\newblock Equivariant mesh attention networks.
\newblock \emph{arXiv preprint arXiv:2205.10662}, 2022.

\bibitem[Wang et~al.(2022)Wang, Walters, and Yu]{wang2022approximately}
Rui Wang, Robin Walters, and Rose Yu.
\newblock Approximately equivariant networks for imperfectly symmetric dynamics.
\newblock In \emph{International Conference on Machine Learning}, pages 23078--23091. PMLR, 2022.

\bibitem[Fu et~al.(2022)Fu, Xie, Rebello, Olsen, and Jaakkola]{fu2022simulate}
Xiang Fu, Tian Xie, Nathan~J Rebello, Bradley~D Olsen, and Tommi Jaakkola.
\newblock Simulate time-integrated coarse-grained molecular dynamics with geometric machine learning.
\newblock \emph{arXiv preprint arXiv:2204.10348}, 2022.

\bibitem[Jaini et~al.(2021)Jaini, Holdijk, and Welling]{jaini2021learning}
Priyank Jaini, Lars Holdijk, and Max Welling.
\newblock Learning equivariant energy based models with equivariant stein variational gradient descent.
\newblock \emph{Advances in Neural Information Processing Systems}, 34:\penalty0 16727--16737, 2021.

\bibitem[Puny et~al.(2021)Puny, Atzmon, Ben-Hamu, Misra, Grover, Smith, and Lipman]{puny2021frame}
Omri Puny, Matan Atzmon, Heli Ben-Hamu, Ishan Misra, Aditya Grover, Edward~J Smith, and Yaron Lipman.
\newblock Frame averaging for invariant and equivariant network design.
\newblock \emph{arXiv preprint arXiv:2110.03336}, 2021.

\bibitem[Gilmer et~al.(2017)Gilmer, Schoenholz, Riley, Vinyals, and Dahl]{gilmer2017neural}
Justin Gilmer, Samuel~S Schoenholz, Patrick~F Riley, Oriol Vinyals, and George~E Dahl.
\newblock Neural message passing for quantum chemistry.
\newblock In \emph{International conference on machine learning}, pages 1263--1272. PMLR, 2017.

\bibitem[Zhang and Chen(2018)]{zhang2018link}
Muhan Zhang and Yixin Chen.
\newblock Link prediction based on graph neural networks.
\newblock \emph{Advances in neural information processing systems}, 31, 2018.

\bibitem[Wang et~al.(2023)Wang, Yang, and Zhang]{wang2023neural}
Xiyuan Wang, Haotong Yang, and Muhan Zhang.
\newblock Neural common neighbor with completion for link prediction.
\newblock \emph{arXiv preprint arXiv:2302.00890}, 2023.

\bibitem[Wu et~al.(2019)Wu, Souza, Zhang, Fifty, Yu, and Weinberger]{wu2019simplifying}
Felix Wu, Amauri Souza, Tianyi Zhang, Christopher Fifty, Tao Yu, and Kilian Weinberger.
\newblock Simplifying graph convolutional networks.
\newblock In \emph{International conference on machine learning}, pages 6861--6871. PMLR, 2019.

\bibitem[Kipf and Welling(2016{\natexlab{a}})]{kipf2016semi}
Thomas~N Kipf and Max Welling.
\newblock Semi-supervised classification with graph convolutional networks.
\newblock \emph{arXiv preprint arXiv:1609.02907}, 2016{\natexlab{a}}.

\bibitem[Hamilton et~al.(2017)Hamilton, Ying, and Leskovec]{hamilton2017inductive}
Will Hamilton, Zhitao Ying, and Jure Leskovec.
\newblock Inductive representation learning on large graphs.
\newblock \emph{Advances in neural information processing systems}, 30, 2017.

\bibitem[Morris et~al.(2019)Morris, Ritzert, Fey, Hamilton, Lenssen, Rattan, and Grohe]{morris2019weisfeiler}
Christopher Morris, Martin Ritzert, Matthias Fey, William~L Hamilton, Jan~Eric Lenssen, Gaurav Rattan, and Martin Grohe.
\newblock Weisfeiler and leman go neural: Higher-order graph neural networks.
\newblock In \emph{Proceedings of the AAAI conference on artificial intelligence}, volume~33, pages 4602--4609, 2019.

\bibitem[Xu et~al.(2018)Xu, Hu, Leskovec, and Jegelka]{xu2018powerful}
Keyulu Xu, Weihua Hu, Jure Leskovec, and Stefanie Jegelka.
\newblock How powerful are graph neural networks?
\newblock \emph{arXiv preprint arXiv:1810.00826}, 2018.

\bibitem[Zhou et~al.(2022)Zhou, Kutyniok, and Ribeiro]{zhou2022ood}
Yangze Zhou, Gitta Kutyniok, and Bruno Ribeiro.
\newblock Ood link prediction generalization capabilities of message-passing gnns in larger test graphs.
\newblock \emph{Advances in Neural Information Processing Systems}, 35:\penalty0 20257--20272, 2022.

\bibitem[Chamberlain et~al.(2022)Chamberlain, Shirobokov, Rossi, Frasca, Markovich, Hammerla, Bronstein, and Hansmire]{chamberlain2022graph}
Benjamin~Paul Chamberlain, Sergey Shirobokov, Emanuele Rossi, Fabrizio Frasca, Thomas Markovich, Nils Hammerla, Michael~M Bronstein, and Max Hansmire.
\newblock Graph neural networks for link prediction with subgraph sketching.
\newblock \emph{arXiv preprint arXiv:2209.15486}, 2022.

\bibitem[Mart{\'\i}nez et~al.(2016)Mart{\'\i}nez, Berzal, and Cubero]{martinez2016survey}
V{\'\i}ctor Mart{\'\i}nez, Fernando Berzal, and Juan-Carlos Cubero.
\newblock A survey of link prediction in complex networks.
\newblock \emph{ACM computing surveys (CSUR)}, 49\penalty0 (4):\penalty0 1--33, 2016.

\bibitem[Kumar et~al.(2020)Kumar, Singh, Singh, and Biswas]{kumar2020link}
Ajay Kumar, Shashank~Sheshar Singh, Kuldeep Singh, and Bhaskar Biswas.
\newblock Link prediction techniques, applications, and performance: A survey.
\newblock \emph{Physica A: Statistical Mechanics and its Applications}, 553:\penalty0 124289, 2020.

\bibitem[Zhu et~al.(2021)Zhu, Zhang, Xhonneux, and Tang]{zhu2021neural}
Zhaocheng Zhu, Zuobai Zhang, Louis-Pascal Xhonneux, and Jian Tang.
\newblock Neural bellman-ford networks: A general graph neural network framework for link prediction.
\newblock \emph{Advances in Neural Information Processing Systems}, 34:\penalty0 29476--29490, 2021.

\bibitem[Huang et~al.(2022)Huang, Chen, Fang, Menkovski, Zhao, Yin, Pei, Mocanu, Wang, Pechenizkiy, et~al.]{huang2022you}
Tianjin Huang, Tianlong Chen, Meng Fang, Vlado Menkovski, Jiaxu Zhao, Lu~Yin, Yulong Pei, Decebal~Constantin Mocanu, Zhangyang Wang, Mykola Pechenizkiy, et~al.
\newblock You can have better graph neural networks by not training weights at all: Finding untrained gnns tickets.
\newblock In \emph{Learning on Graphs Conference}, pages 8--1. PMLR, 2022.

\bibitem[B{\"o}ker et~al.(2023)B{\"o}ker, Levie, Huang, Villar, and Morris]{boker2023fine}
Jan B{\"o}ker, Ron Levie, Ningyuan Huang, Soledad Villar, and Christopher Morris.
\newblock Fine-grained expressivity of graph neural networks.
\newblock \emph{arXiv preprint arXiv:2306.03698}, 2023.

\bibitem[Dong et~al.(2023)Dong, Guo, and Chawla]{dong2023pure}
Kaiwen Dong, Zhichun Guo, and Nitesh~V Chawla.
\newblock Pure message passing can estimate common neighbor for link prediction.
\newblock \emph{arXiv preprint arXiv:2309.00976}, 2023.

\bibitem[Dong et~al.(2024)Dong, Guo, and Chawla]{dong2024you}
Kaiwen Dong, Zhichun Guo, and Nitesh~V Chawla.
\newblock You do not have to train graph neural networks at all on text-attributed graphs.
\newblock \emph{arXiv preprint arXiv:2404.11019}, 2024.

\bibitem[Sato(2024)]{sato2024training}
Ryoma Sato.
\newblock Training-free graph neural networks and the power of labels as features.
\newblock \emph{arXiv preprint arXiv:2404.19288}, 2024.

\bibitem[Kipf and Welling(2016{\natexlab{b}})]{kipf2016variational}
Thomas~N Kipf and Max Welling.
\newblock Variational graph auto-encoders.
\newblock \emph{arXiv preprint arXiv:1611.07308}, 2016{\natexlab{b}}.

\bibitem[Yun et~al.(2021)Yun, Kim, Lee, Kang, and Kim]{yun2021neo}
Seongjun Yun, Seoyoon Kim, Junhyun Lee, Jaewoo Kang, and Hyunwoo~J Kim.
\newblock Neo-gnns: Neighborhood overlap-aware graph neural networks for link prediction.
\newblock \emph{Advances in Neural Information Processing Systems}, 34:\penalty0 13683--13694, 2021.

\bibitem[Pan et~al.(2021)Pan, Shi, and Dokmani{\'c}]{pan2021neural}
Liming Pan, Cheng Shi, and Ivan Dokmani{\'c}.
\newblock Neural link prediction with walk pooling.
\newblock \emph{arXiv preprint arXiv:2110.04375}, 2021.

\bibitem[Sato et~al.(2021)Sato, Yamada, and Kashima]{sato2021random}
Ryoma Sato, Makoto Yamada, and Hisashi Kashima.
\newblock Random features strengthen graph neural networks.
\newblock In \emph{Proceedings of the 2021 SIAM international conference on data mining (SDM)}, pages 333--341. SIAM, 2021.

\bibitem[Abboud et~al.(2020)Abboud, Ceylan, Grohe, and Lukasiewicz]{abboud2020surprising}
Ralph Abboud, Ismail~Ilkan Ceylan, Martin Grohe, and Thomas Lukasiewicz.
\newblock The surprising power of graph neural networks with random node initialization.
\newblock \emph{arXiv preprint arXiv:2010.01179}, 2020.

\bibitem[Cui et~al.(2022)Cui, Lu, Li, and Yang]{cui2022positional}
Hejie Cui, Zijie Lu, Pan Li, and Carl Yang.
\newblock On positional and structural node features for graph neural networks on non-attributed graphs.
\newblock In \emph{Proceedings of the 31st ACM International Conference on Information \& Knowledge Management}, pages 3898--3902, 2022.

\bibitem[Lin et~al.(2020)Lin, He, Zhang, Cheng, and Cai]{lin2020initialization}
Wenqing Lin, Feng He, Faqiang Zhang, Xu~Cheng, and Hongyun Cai.
\newblock Initialization for network embedding: A graph partition approach.
\newblock In \emph{Proceedings of the 13th International Conference on Web Search and Data Mining}, pages 367--374, 2020.

\bibitem[Kanerva(2009)]{kanerva2009hyperdimensional}
Pentti Kanerva.
\newblock Hyperdimensional computing: An introduction to computing in distributed representation with high-dimensional random vectors.
\newblock \emph{Cognitive computation}, 1:\penalty0 139--159, 2009.

\bibitem[Kanerva(2018)]{kanerva2018computing}
Pentti Kanerva.
\newblock Computing with high-dimensional vectors.
\newblock \emph{IEEE Design \& Test}, 36\penalty0 (3):\penalty0 7--14, 2018.

\bibitem[Morris and Ningyuan~Huang(2023)]{finegranedmorris}
Christopher Morris and Teresa Ningyuan~Huang.
\newblock Gin implementation.
\newblock \url{https://github.com/nhuang37/finegrain_expressivity_GNN/blame/main/GNN_untrained/gnn_baselines/gnn_architectures.py}, 2023.
\newblock Accessed: 2023-11-21.

\bibitem[Eaton(2007)]{eaton2007random}
Morris~L Eaton.
\newblock Random vectors.
\newblock In \emph{Multivariate Statistics}, volume~53, pages 70--103. Institute of Mathematical Statistics, 2007.

\bibitem[Nt and Maehara(2019)]{nt2019revisiting}
Hoang Nt and Takanori Maehara.
\newblock Revisiting graph neural networks: All we have is low-pass filters.
\newblock \emph{arXiv preprint arXiv:1905.09550}, 2019.

\bibitem[Rapoport(1953)]{rapoport1953spread}
Anatol Rapoport.
\newblock Spread of information through a population with socio-structural bias: I. assumption of transitivity.
\newblock \emph{The bulletin of mathematical biophysics}, 15:\penalty0 523--533, 1953.

\bibitem[Holland and Leinhardt(1971)]{holland1971transitivity}
Paul~W Holland and Samuel Leinhardt.
\newblock Transitivity in structural models of small groups.
\newblock \emph{Comparative group studies}, 2\penalty0 (2):\penalty0 107--124, 1971.

\bibitem[Lü and Zhou(2011)]{LU20111150}
Linyuan Lü and Tao Zhou.
\newblock Link prediction in complex networks: A survey.
\newblock \emph{Physica A: Statistical Mechanics and its Applications}, 390\penalty0 (6):\penalty0 1150--1170, 2011.
\newblock ISSN 0378-4371.
\newblock \doi{https://doi.org/10.1016/j.physa.2010.11.027}.
\newblock URL \url{https://www.sciencedirect.com/science/article/pii/S037843711000991X}.

\bibitem[Adamic and Adar(2003)]{adamic2003friends}
Lada~A Adamic and Eytan Adar.
\newblock Friends and neighbors on the web.
\newblock \emph{Social networks}, 25\penalty0 (3):\penalty0 211--230, 2003.

\bibitem[Zhou et~al.(2009)Zhou, L{\"u}, and Zhang]{zhou2009predicting}
Tao Zhou, Linyuan L{\"u}, and Yi-Cheng Zhang.
\newblock Predicting missing links via local information.
\newblock \emph{The European Physical Journal B}, 71:\penalty0 623--630, 2009.

\bibitem[Katz(1953)]{katz1953new}
Leo Katz.
\newblock A new status index derived from sociometric analysis.
\newblock \emph{Psychometrika}, 18\penalty0 (1):\penalty0 39--43, 1953.

\bibitem[Brin and Page(1998)]{brin1998anatomy}
Sergey Brin and Lawrence Page.
\newblock The anatomy of a large-scale hypertextual web search engine.
\newblock \emph{Computer networks and ISDN systems}, 30\penalty0 (1-7):\penalty0 107--117, 1998.

\bibitem[Jeh and Widom(2002)]{jeh2002simrank}
Glen Jeh and Jennifer Widom.
\newblock Simrank: a measure of structural-context similarity.
\newblock In \emph{Proceedings of the eighth ACM SIGKDD international conference on Knowledge discovery and data mining}, pages 538--543, 2002.

\bibitem[Li et~al.(2023)Li, Shomer, Mao, Zeng, Ma, Shah, Tang, and Yin]{li2023evaluating}
Juanhui Li, Harry Shomer, Haitao Mao, Shenglai Zeng, Yao Ma, Neil Shah, Jiliang Tang, and Dawei Yin.
\newblock Evaluating graph neural networks for link prediction: Current pitfalls and new benchmarking.
\newblock \emph{Advances in Neural Information Processing Systems}, 36:\penalty0 3853--3866, 2023.

\bibitem[Yang et~al.(2016)Yang, Cohen, and Salakhudinov]{yang2016revisiting}
Zhilin Yang, William Cohen, and Ruslan Salakhudinov.
\newblock Revisiting semi-supervised learning with graph embeddings.
\newblock In \emph{International conference on machine learning}, pages 40--48. PMLR, 2016.

\bibitem[Bojchevski and G{\"u}nnemann(2017)]{bojchevski2017deep}
Aleksandar Bojchevski and Stephan G{\"u}nnemann.
\newblock Deep gaussian embedding of graphs: Unsupervised inductive learning via ranking.
\newblock \emph{arXiv preprint arXiv:1707.03815}, 2017.

\bibitem[Hu et~al.(2020)Hu, Fey, Zitnik, Dong, Ren, Liu, Catasta, and Leskovec]{hu2020ogb}
Weihua Hu, Matthias Fey, Marinka Zitnik, Yuxiao Dong, Hongyu Ren, Bowen Liu, Michele Catasta, and Jure Leskovec.
\newblock Open graph benchmark: Datasets for machine learning on graphs.
\newblock \emph{arXiv preprint arXiv:2005.00687}, 2020.

\bibitem[Newman(2006)]{newman2006finding}
Mark~EJ Newman.
\newblock Finding community structure in networks using the eigenvectors of matrices.
\newblock \emph{Physical review E}, 74\penalty0 (3):\penalty0 036104, 2006.

\bibitem[Watts and Strogatz(1998)]{watts1998collective}
Duncan~J Watts and Steven~H Strogatz.
\newblock Collective dynamics of ‘small-world’networks.
\newblock \emph{Nature}, 393\penalty0 (6684):\penalty0 440--442, 1998.

\bibitem[Ackland et~al.(2005)]{ackland2005mapping}
Robert Ackland et~al.
\newblock Mapping the us political blogosphere: Are conservative bloggers more prominent?
\newblock In \emph{BlogTalk Downunder 2005 Conference, Sydney}. BlogTalk Downunder 2005 Conference, Sydney, 2005.

\bibitem[Spring et~al.(2004)Spring, Mahajan, Wetherall, and Anderson]{spring2004measuring}
Neil Spring, Ratul Mahajan, David Wetherall, and Thomas Anderson.
\newblock Measuring isp topologies with rocketfuel.
\newblock \emph{IEEE/ACM Transactions on networking}, 12\penalty0 (1):\penalty0 2--16, 2004.

\bibitem[Batagelj and Mrvar(2006)]{usair}
Vladimir Batagelj and Andrej Mrvar.
\newblock Usair data.
\newblock \url{http://vlado.fmf.uni-lj.si/pub/networks/data/}, 2006.
\newblock Accessed: 27-01-2024.

\bibitem[Von~Mering et~al.(2002)Von~Mering, Krause, Snel, Cornell, Oliver, Fields, and Bork]{von2002comparative}
Christian Von~Mering, Roland Krause, Berend Snel, Michael Cornell, Stephen~G Oliver, Stanley Fields, and Peer Bork.
\newblock Comparative assessment of large-scale data sets of protein--protein interactions.
\newblock \emph{Nature}, 417\penalty0 (6887):\penalty0 399--403, 2002.

\bibitem[Zhang et~al.(2018)Zhang, Cui, Jiang, and Chen]{zhang2018beyond}
Muhan Zhang, Zhicheng Cui, Shali Jiang, and Yixin Chen.
\newblock Beyond link prediction: Predicting hyperlinks in adjacency space.
\newblock In \emph{Proceedings of the AAAI Conference on Artificial Intelligence}, volume 32-1, 2018.

\bibitem[Kingma and Ba(2014)]{kingma2014adam}
Diederik~P Kingma and Jimmy Ba.
\newblock Adam: A method for stochastic optimization.
\newblock \emph{arXiv preprint arXiv:1412.6980}, 2014.

\bibitem[pyGteam(2021)]{pyglinkpred}
pyGteam.
\newblock Link prediction on pyg.
\newblock \url{https://github.com/pyg-team/pytorch_geometric/blob/master/examples/link_pred.py}, 2021.
\newblock Accessed: 2023-11-21.

\end{thebibliography}

\appendix

\section{Dataset details}
\label{section: Data}
Summary statistics and sources of data sets are given in Table \ref{table: dataset description}. 
\begin{table}[!ht]
\centering 
\caption{Overview of the datasets, sources, and node features for attributed graphs (top group) used in our experimental evaluation.}
\label{table: dataset description}
\begin{tabular}{lccccc}
\textbf{Dataset }      & \textbf{$\vert$V$\vert$}             & \textbf{$\vert$E$\vert$  }              & \textbf{Features}                     \\ \hline \hline
\textbf{Cora small \cite{yang2016revisiting}}                &    2,708         &     10,556          &  1,433     \\
\textbf{CiteSeer small \cite{yang2016revisiting}}                &        3,327     &      9,104         &    3,703     \\
\textbf{Cora  \cite{bojchevski2017deep}}                & 19,793            & 126,842              & 8,710      \\
\textbf{Cora ML \cite{bojchevski2017deep}}                & 2,995            & 16,316              & 2,879      \\
\textbf{PubMed \cite{bojchevski2017deep}}             & 19,717           & 88,648             & 500      \\
\textbf{CiteSeer  \cite{bojchevski2017deep}}            & 4,230            & 10,674              & 602      \\
\textbf{DBLP \cite{bojchevski2017deep}}      & 17,716        & 105,734          & 1,639       \\
\hline
\textbf{OGBL-Collab \cite{hu2020ogb}}      &   235,868     &     1,285,465      &   128      \\
\textbf{OGBL-PPA \cite{hu2020ogb}}      &   576,289	     &     30,326,273	      &    58    \\ 
\textbf{OGBL-DDI \cite{hu2020ogb}}      &   4,267     &     1,334,889      &    -     \\ 
\hline

\textbf{NS \cite{newman2006finding}}      &    1,461     &    2,742       &    -    \\
\textbf{Celegans \cite{watts1998collective}}      &   297      &     2,148      &    -    \\
\textbf{PB \cite{ackland2005mapping}}      &    1,222     &    16,714       &    -    \\
\textbf{Power \cite{watts1998collective}}      &    4,941     &      6,594      &    -    \\
\textbf{Router \cite{spring2004measuring}}      &    5,022     &     6,258      &    -    \\
\textbf{USAir \cite{usair}}      &    332     &    2,126       &    -    \\
\textbf{Yeast \cite{von2002comparative}}      &    2,375     &    11,693       &    -    \\
\textbf{E-coli \cite{zhang2018beyond}}      &    1,805     &       15,660    &    -    \\
\hline \hline
\end{tabular}%

\end{table}

\section{Additional Results}
In this section, we present additional results for the attributed datasets measured using ROC-AUC, and we also include the results for the OGB datasets.

\begin{table}[!htbp]
\centering 
\caption{Link Prediction accuracy for attributed networks as measured by ROC-AUC. Red values correspond to the overall best model for each dataset, and blue values indicate the best-performing model within the same category of message passing layers.}
\label{table: results roc}
\resizebox{\textwidth}{!}{%
\begin{tabular}{lccccccc}
\textbf{Models}& \textbf{Cora (small)} & \textbf{CiteSeer (small) } & \textbf{Cora } & \textbf{Cora ML} & \textbf{PubMed} & \textbf{CiteSeer}    & \textbf{DBLP}  \\ \hline  \hline& \textit{ROC-AUC} & \textit{ROC-AUC} & \textit{ROC-AUC} & \textit{ROC-AUC} & \textit{ROC-AUC} & \textit{ROC-AUC}    & \textit{ROC-AUC} \\ \hline  \hline
GCN &92.82 ± 0.83&91.67 ± 1.14&97.87 ± 0.18&94.67 ± 0.51&97.54 ± 0.09&94.22 ± 0.77&96.63 ± 0.12   \\
SGCN &\textcolor{red}{95.3 ± 0.68}&\textcolor{red}{96.22 ± 0.4}&\textcolor{red}{98.6 ± 0.07}&\textcolor{red}{96.82 ± 0.43}&\textcolor{red}{97.94 ± 0.16}&\textcolor{red}{95.9 ± 0.77}&\textcolor{red}{97.1 ± 0.14}    \\ 
UTGCN &93.82 ± 0.68&96.0 ± 0.24&95.72 ± 0.12&93.93 ± 0.45&94.29 ± 0.24&92.74 ± 0.81&94.91 ± 0.32   \\ \hline
SAGE &91.41 ± 0.38&90.78 ± 1.79&97.7 ± 0.08&94.38 ± 0.58&95.6 ± 0.18&93.58 ± 0.87&96.16 ± 0.25   \\
SSAGE &\textcolor{blue}{94.47 ± 0.64}&95.75 ± 0.29&\textcolor{blue}{98.23 ± 0.1}&\textcolor{blue}{95.74 ± 0.6}&\textcolor{blue}{95.88 ± 0.22}&\textcolor{blue}{95.14 ± 0.9}&\textcolor{blue}{96.29 ± 0.12}   \\  
UTSAGE &92.77 ± 0.5&\textcolor{blue}{96.07 ± 0.43}&96.85 ± 0.13&93.45 ± 0.81&88.12 ± 0.29&91.78 ± 0.85&93.36 ± 0.31   \\ \hline
GIN &91.65 ± 0.73&90.62 ± 1.17&97.69 ± 0.13&94.54 ± 0.32&96.27 ± 0.13&92.88 ± 0.87&\textcolor{blue}{96.11 ± 0.25}   \\ 
GraphConv &92.06 ± 0.67&91.24 ± 0.32&\textcolor{blue}{97.94 ± 0.11}&\textcolor{blue}{95.34 ± 0.25}&\textcolor{blue}{96.39 ± 0.15}&92.7 ± 0.73&96.09 ± 0.23   \\ \hdashline
SGIN &\textcolor{blue}{92.87 ± 0.37}&\textcolor{blue}{93.6 ± 0.48}&97.82 ± 0.11&95.26 ± 0.41&96.37 ± 0.23&\textcolor{blue}{94.13 ± 0.4}&95.85 ± 0.11   \\ 
UTGIN &85.45 ± 1.28&85.7 ± 0.82&88.93 ± 0.35&86.86 ± 0.98&88.76 ± 0.25&91.11 ± 0.93&92.25 ± 0.29 
   \\ 
   \hline \hline

\end{tabular}%
}
\end{table}

\begin{table}[!htbp]
\centering 
\caption{Link Prediction accuracy for OGB datasets. Red values correspond to the overall best model for each dataset, and blue values indicate the best-performing model within the same category of message passing layers.}
\label{table: resultsogb}

\begin{tabular}{lccc}
\textbf{Models}& \textbf{ ogbl-collab } & \textbf{ ogbl-ppa } & \textbf{ ogbl-ddi }    \\ \hline  \hline
& \textit{Hits@50} & \textit{Hits@100} & \textit{Hits@20}   \\ \hline  \hline
GCN &\textcolor{blue}{55.58 ± 3.84}&\textcolor{blue}{30.84 ± 1.78}&\textcolor{blue}{48.58 ± 7.11} \\
SGCN &54.13 ± 0.97&12.34 ± 1.95&22.14 ± 3.38 \\ \hline
SAGE &45.90 ± 8.67&\textcolor{blue}{23.88 ± 1.63}&\textcolor{blue}{22.00 ± 14.54}  \\
SSAGE &\textcolor{blue}{49.92 ± 1.52}&8.86 ± 0.80&19.44 ± 10.84  \\ \hline
NCNC (GCN) &62.29 ± 3.34&OOM&\textcolor{blue}{46.53 ± 29.24}  \\
SNCNC (SGCN) &\textcolor{blue}{65.40 ± 0.46}&\textcolor{blue}{46.02 ± 1.19}&13.55 ± 13.48  \\\hline
NCNC (SAGE) &\textcolor{blue}{64.91 ± 1.50}&\textcolor{blue}{55.60 ± 3.06}&\textcolor{blue}{40.38 ± 19.71}  \\
SNCNC (SSAGE) &62.31 ± 1.90&38.17 ± 2.37&13.19 ± 9.33 \\\hline
NCNC (GIN) &\textcolor{red}{65.87 ± 0.74}&\textcolor{red}{59.47 ± 1.60}&32.62 ± 30.71  \\
SNCNC (SGIN) &16.69 ± 2.43&21.81 ± 0.75&\textcolor{red}{54.53 ± 11.02} \\
\hline \hline

\end{tabular}%

\end{table}

\section{Experimental setup and Hyperparameter choices}
\label{section: Hyperparameter choices}
For each model, the optimal values of the learning rate, the number of layers, and hidden dimensions are determined through an exhaustive search over the values given in Table \ref{table: hyperparameters}). The optimal hyperparameters values for attributed and non-attributed datasets are given in Table \ref{tab:grid choices} and Table \ref{tab:grid choices noattr}, respectively. We implement a three-fold cross-validation procedure to select the optimal hyperparameter values.

We use Adam \cite{kingma2014adam} as an optimization function and employ binary cross entropy with logits as our loss function. All datasets are preprocessed by normalizing the node features and randomly splitting them. For non-attributed datasets, $10\%$ of the data is allocated to the test set, $5\%$ to the validation set \cite{zhang2018link}, and the remaining data is used for the training set. In contrast, for attributed datasets, the split is $20\%$ for the test set, $10\%$ for the validation set, and the remainder for the training set \cite{wang2023neural}.Each model configuration is run $10$ times, with the results averaged over these runs. 
Our training and testing procedures are based on the methodology outlined in \cite{pyglinkpred}, where we perform a new round of negative edge sampling for each training epoch. We limit the maximum number of epochs to 10,000 and also incorporate an early stopping mechanism in our training process by terminating training whenever there is no improvement in the validation set results over a span of $250$ epochs.

All hyperparameter searches and experiments were conducted on a workstation with AMD Ryzen Threadripper PRO 5965WX 24-Cores with 256 GB of memory and two Nvidia GeForce RTX 3090 Super GPU, and also AMD Ryzen 9 7900X 12-Cores with 64 GB of memory and an Nvidia GeForce RTX 4080 GPU.

\begin{table}[!htb]
\centering
\caption{The hyperparameter space for our experiments. It is worth noting that only the number of MPNN layers applies to the untrained models.}
\label{table: hyperparameters}
\begin{tabular}{lc}
\textbf{Hyperparameter}  & \textbf{Values}                                   \\ \hline \hline
Number of MPNN layers & 1,2,3                               \\
Learning Rate         & 0.2, 0.1,0.01, 0.001, 0.0001 \\
Hidden Dimensions     & 16, 64, 128             \\  \hline
\end{tabular}

\end{table}

\begin{table}[!htbp]
    \centering
        \caption{Optimal hyperparameter values for attributed datasets (MaxEpochs=10,000).}
    \label{tab:grid choices}
\resizebox{\textwidth}{!}{%
\begin{tabular}{l|ccc|ccc|ccc|ccc|ccc|ccc|ccc}
\multirow{2}{*}{} &\multicolumn{3}{c}{\textbf{Cora small}} & \multicolumn{3}{c}{\textbf{CiteSeer small}} & \multicolumn{3}{c}{\textbf{Cora}} & \multicolumn{3}{c}{\textbf{Cora ML}} & \multicolumn{3}{c}{\textbf{PubMed}} & \multicolumn{3}{c}{\textbf{CiteSeer}} & \multicolumn{3}{c}{\textbf{DBLP}}  \\ \hline \hline
                  & lr.       & hd.      & nl.      & lr.    & hd.    & nl.    & lr.     & hd.     & nl.    & lr.      & hd.     & nl.     & lr.       & hd.      & nl.       & lr.       & hd.      & nl.   & lr.       & hd.      & nl.        \\ \hline \hline
GCN       &     0.001      &    128     &     1     &   0.001    &    128    &     1    &     0.001      &    64     &    1     &    0.001   &   64     &  1      &   0.01      &    64     &    1   &   0.01      &    64     &    1     &      0.001     &    128      &    1                           \\
SGCN   &      0.001     &      64   &    1      &   0.001    &    128    &     1     &    0.001       &   128      &     1     &   0.001    &   128     &  2      &    0.001     &    128     &   1    &    0.001     &    128     &   2      &     0.2      &   128       & 2                            \\
UTGCN    &           &         &      2    &       &        &    2    &           &         &      2    &       &        &   2     &         &         &   2    &         &         &    3     &           &          &       2                    \\ \hline 
SAGE    &     0.01      &     128    &      1    &   0.01    &    16    &    1      &    0.01       &   128      &     1     &    0.001   &    128    &    1    &    0.01     &    128     &    1   &    0.001     &     128    &    1     &      0.001     &     64     &    1                     \\
SSAGE    &      0.0001     &    128     &     1     &     0.0001  &   128     &   2   &     0.1      &    64     &     1     &    0.001   &   64     &   1     &     0.001    &     128    &   2    &      0.01   &    128     &   3      &     0.01      &     64     &   2                  \\
UTSAGE   &           &         &      2    &       &        &    2    &           &         &     2     &       &        &    2    &         &         &    2   &         &         &   2      &           &          &        2                        \\ \hline 
GIN       &     0.001      &     128    &     1     &   0.001    &   128     &    1      &     0.001      &    128     &     1     &   0.001    &   128     &    1    &     0.001    &    128     &   1    &   0.001      &    128     &    1     &    0.001       &     128     &        1                  \\
GraphConv    &    0.0001       &    64     &     1     &    0.0001   &    128    &   1     &     0.001      &    64     &    1      &   0.0001    &   128     &   1     &   0.0001      &    128     &   1    &     0.001    &    128     &    1     &     0.001      &     128     &        1                       \\ \hdashline
SGIN    &    0.001       &    64     &      1    &   0.0001    &   128     &     2    &       0.0001    &   128      &    1      &   0.001    &    64    &   1     &     0.01    &   128      &    1   &    0.0001     &    128     &    1     &     0.001      &     128     &   1                   \\
UTGIN    &           &         &      1    &       &        &   1  &           &         &     1     &       &        &    1    &         &         &     1  &         &         &      1   &           &          &         1              \\ \hline \hline
\end{tabular}%
}

\end{table}

\begin{table}[!htbp]
    \centering
        \caption{Hyperparameter choices for each model in each of the non-attributed  dataset.}
    \label{tab:grid choices noattr}
\resizebox{\textwidth}{!}{%
\begin{tabular}{l|ccc|ccc|ccc|ccc|ccc|ccc|ccc|ccc}
\multirow{2}{*}{} & \multicolumn{3}{c}{\textbf{NS}} & \multicolumn{3}{c}{\textbf{Celegans}} & \multicolumn{3}{c}{\textbf{PB}} & \multicolumn{3}{c}{\textbf{Power}} & \multicolumn{3}{c}{\textbf{Router}} & \multicolumn{3}{c}{\textbf{USAir}} & \multicolumn{3}{c}{\textbf{Yeast}} & \multicolumn{3}{c}{\textbf{E-coli}} \\ \hline \hline
                  & lr.       & hd.      & nl.      & lr.    & hd.    & nl.    & lr.     & hd.     & nl.    & lr.      & hd.     & nl.     & lr.       & hd.      & nl.     & lr.      & hd.     & nl.     & lr.       & hd.      & nl.   & lr.       & hd.      & nl.    \\ \hline \hline
GCN           &     0.01      &    64     &    3      &   0.001    &   128     &    1    &     0.01    &    128     &    2   &     0.001    &   64      &   3      &     0.2      &    128      &   3    &    0.001     &    128     &    2     &      0.01     &    64      &      3             &      0.1     &    128      &      1             \\
SGCN        &    0.1       &    128     &     3     &   0.01    &    128    &    2    &     0.1    &    128     &   2    &      0.001   &    128     &   3      &     0.2      &    64      &    3   &  0.1       &    64     &   2      &     0.01      &   128       &     2              &     0.01      &    16      &        1           \\
UTGCN        &           &         &     3     &       &        &    2    &         &         &  2     &         &         &    3     &           &          &    2   &         &         &    2     &           &          &      2             &           &          &     2              \\ \hline 
SAGE          &    0.01       &    64     &     2     &  0.01     &    128    &   2     &     0.01    &     128    &    2   &      0.01   &    64     &    3     &      0.2     &    16      &   1    &     0.01    &    64     &    1     &     0.01      &    64      &      2             &     0.01      &     128     &      1             \\
SSAGE       &       0.01    &    128     &    1      &    0.01   &   16     &    2    &     0.01    &    128     &   1    &     0.001    &    128     &   3      &     0.001      &     64     &   3    &    0.1     &    64     &    2     &     0.001      &     128     &       1            &     0.01      &     128     &     1              \\
UTSAGE       &           &         &     2     &       &        &    2    &         &         &  2     &         &         &     3    &           &          &     2  &         &         &     2    &           &          &    2               &           &          &      2             \\ \hline 
GIN           &      0.001     &    128     &    3      &   0.001    &   64     &    1    &    0.001     &    128     &   1    &     0.01    &    128     &     3    &     0.1      &    128      &    2   &    0.0001     &    128     &    2     &      0.001     &    128      &      1             &     0.01      &     128     &        1           \\ 
GraphConv         &     0.001      &    128     &    1      &   0.0001    &   128     &   1     &   0.0001      &     64    &   1    &     0.0001    &    128     &     3    &      0.001     &   16       &  1     &    0.01     &    64     &   2      &     0.0001      &    64      &      1             &      0.01     &     64     &      1             \\ \hdashline
SGIN        &      0.0001     &    128     &    2      &    0.01   &    128    &    1    &     0.2    &    64     &   1    &    0.0001     &    128     &    2     &      0.0001     &    128      &   1    &    0.1     &    64     &    1     &     0.001      &     128     &       1            &     0.001      &     64     &       1            \\ 
UTGIN      &           &         &     2     &       &        &      1  &         &         &    1   &         &         &      3   &           &          &     2  &         &         &   2      &           &          &     2              &           &          &     1              \\ \hline \hline
\end{tabular}%
}

\end{table}

\section{Runtime Analysis and Training Efficiency}
\label{section: runtime}

\paragraph{Efficiency of SMPNNs}
While we allocated a very generous limit of 10,000 epochs for training models in the main paper to ensure models can reach their best possible performance in order to compare the computational efficiency of the simplified models to their fully trained counterparts we also consider an experimental setting where we restrict the maximum number of training epochs to 100. We find that simplified models achieved convergence even for larger learning rates and considerably faster than their fully trained models. Even when constrained to 100 training epochs simplified models maintain scores that are almost identical to those presented in Table \ref{table: results 100}, while fully trained architectures suffer from the increased learning rates and require in general more epochs to converge. This leads to training efficiency gains similar to those reported by \cite{wu2019simplifying} in the case of node classification. 

In Table \ref{table: results 100}, it is evident that the simplified models consistently outperform the fully trained models across all datasets by a considerable margin.  Furthermore, as demonstrated in Table \ref{table: results}, the fully trained models nearly achieve their peak accuracy within just 100 epochs, indicating that extended training offers minimal additional benefit. This also implies that the Simplified models are more efficient in terms of both time and resources required for training.

The hyperpameter space used for the computational efficiency experiments is the same as in Table\ref{tab:grid choices}, except that we only use 100 epochs. 

\begin{figure}
    \centering
    \includegraphics[width=\textwidth]{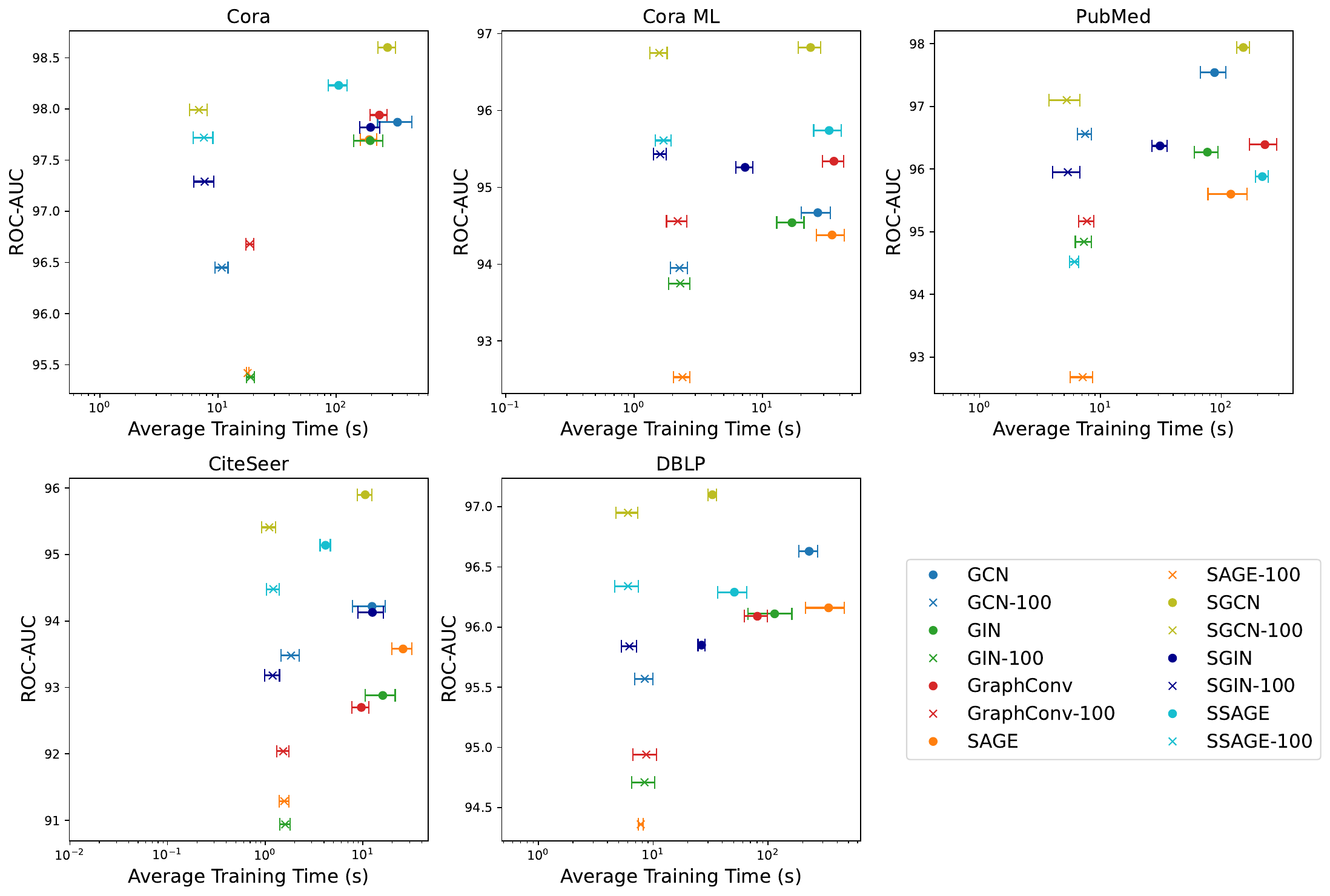}
    \caption{Average runtimes (in seconds) for training and inference for attributed data sets.}
    \label{fig: training times}
\end{figure}

\paragraph{Efficiency of UTMP}
In Figure \ref{fig: training times}, we presented the training times for both simplified and fully trained models. 
The prediction times for UT models are excluded, as they require only a single "epoch" for making the predictions, unlike other methods that necessitate prolonged training periods. This characteristic of UT models leads to a substantial reduction in both time consumption and electricity costs.

Despite a minor trade-off in accuracy on attributed graphs, UT models frequently outperform in terms of accuracy on unattributed graphs across numerous datasets. 
In practical applications, the efficiency of UTMP models could translate to significant savings in energy consumption and hence environmental footprint which can outweigh marginal improvements in accuracy in settings where either computational resources are limited or reducing energy consumption/cost and environmental impact of models take priority. 
This makes UT models particularly appealing for large-scale applications where operational efficiency and cost reduction are critical. Additionally, the societal impact of using UT models includes a lower environmental footprint due to reduced energy consumption, aligning with sustainable and environmentally friendly practices.
\begin{table}[!htbp]
    \centering
        \caption{Optimal hyperparameter values for attributed datasets for MaxEpochs=100.}
    \label{tab:grid choices_100}
\resizebox{\textwidth}{!}{%
\begin{tabular}{l|ccc|ccc|ccc|ccc|ccc|ccc|ccc}
\multirow{2}{*}{} &\multicolumn{3}{c}{\textbf{Cora (small)}} & \multicolumn{3}{c}{\textbf{CiteSeer (small)}} & \multicolumn{3}{c}{\textbf{Cora}} & \multicolumn{3}{c}{\textbf{Cora ML}} & \multicolumn{3}{c}{\textbf{PubMed}} & \multicolumn{3}{c}{\textbf{CiteSeer}} & \multicolumn{3}{c}{\textbf{DBLP}} \\ \hline \hline
                  & lr.       & hd.      & nl.      & lr.    & hd.    & nl.    & lr.     & hd.     & nl.    & lr.      & hd.     & nl.     & lr.       & hd.      & nl.       & lr.       & hd.      & nl.   & lr.       & hd.      & nl.      \\ \hline \hline
GCN                  &    0.01       &     128     &   1    &      0.01       &     128     &   1        &   0.01       &     64     &   1     &     0.01       &     64     &   1     &     0.01       &     64     &   1     &     0.01       &     64     &   1        &   0.01       &     128     &   1         \\
SGCN       &      0.1       &    128     &     1        &  0.1       &    128     &     2      &  0.01          &      128    &     1     &    0.01          &      128    &     1    &    0.01          &      128    &     1    &     0.01          &      64    &     2       &      0.1          &      128    &     2            \\ \hline 
SAGE               &    0.01       &     128     &   1   &     0.01       &     128     &   1       &     0.01       &     64     &   1    &    0.01       &     64     &   1      &    0.01       &     128     &   1    &    0.01       &     128     &   1        &    0.01       &     128     &   1       \\
SSAGE       &   0.2          &    128     &     2        &   0.1          &    128     &     2      &  0.01          &      128    &     1     &    0.01          &      128    &     1    &    0.01          &      128    &     1    &    0.01          &      64    &     1      &      0.1          &      128    &     2             \\ \hline 
GIN       &      0.001       &     128     &   1       &   0.001       &     128     &   1     &     0.001       &   128       &    1      &    0.001       &   128       &    1      &    0.001       &     128     &   1   &     0.01       &     64     &   1      &    0.01       &     64     &   1          \\
GraphConv    &      0.001       &     128     &   1     &     0.001       &     128     &   1     &        0.001       &   128       &    1    &    0.001       &   128       &    1     &     0.001       &     128     &   1     &      0.01       &     128     &   1     &      0.001       &     128     &   1        \\ \hdashline
SGIN    &      0.001     &    128     &     1      &    0.001     &    128     &     1     &     0.001       &     128     &      1    &    0.001          &      128    &     1      &    0.001          &      128    &     1    &    0.001          &      128    &     1       &       0.001          &      128    &     1          \\ \hline \hline
\end{tabular}%
}

\end{table}

\begin{table}[!htbp]
\centering 
\caption{Link Prediction accuracy for attributed networks as measured by ROC-AUC. Red values correspond to the overall best model for each dataset, and blue values indicate the best-performing model within the same category of message passing layers. The models are trained only for MaxEpochs = 100.}
\label{table: results 100}
\resizebox{\textwidth}{!}{%
\begin{tabular}{lccccccc}
\textbf{Models}& \textbf{Cora (small)} & \textbf{CiteSeer (small)} & \textbf{Cora} & \textbf{Cora ML} & \textbf{PubMed} & \textbf{CiteSeer}    & \textbf{DBLP}  \\ \hline  \hline
GCN &91.44 ± 1.31&91.48 ± 0.67&96.45 ± 0.29&93.95 ± 0.54&96.56 ± 0.22&93.48 ± 0.81&95.57 ± 0.18     \\
SGCN &\textcolor{red}{94.58 ± 1.27}&\textcolor{red}{96.4 ± 0.97}&\textcolor{red}{97.99 ± 0.06}&\textcolor{red}{96.75 ± 0.3}&\textcolor{red}{97.1 ± 0.17}&\textcolor{red}{95.41 ± 0.76}&\textcolor{red}{96.95 ± 0.1}      \\ \hline
SAGE &90.2 ± 1.67&90.34 ± 1.87&95.42 ± 0.22&92.53 ± 0.69&92.68 ± 0.5&91.29 ± 1.32&94.36 ± 0.32     \\
SSAGE &\textcolor{blue}{93.98 ± 1.08}&\textcolor{blue}{95.77 ± 1.02}&\textcolor{blue}{97.72 ± 0.08}&\textcolor{blue}{95.61 ± 0.38}&\textcolor{blue}{94.52 ± 0.18}&\textcolor{blue}{94.48 ± 0.96}&\textcolor{blue}{96.34 ± 0.12}     \\   \hline
GIN &90.39 ± 0.6&88.27 ± 0.61&95.38 ± 0.29&93.75 ± 0.24&94.84 ± 0.28&90.94 ± 0.72&94.71 ± 0.26     \\
GraphConv &91.57 ± 1.33&90.79 ± 0.91&96.68 ± 0.16&94.56 ± 0.48&95.17 ± 0.3&92.04 ± 0.96&94.94 ± 0.11     \\ 
SGIN &\textcolor{blue}{92.72 ± 1.23}&\textcolor{blue}{93.11 ± 0.25}&\textcolor{blue}{97.29 ± 0.08}&\textcolor{blue}{95.43 ± 0.27}&\textcolor{blue}{95.95 ± 0.21}&\textcolor{blue}{93.18 ± 0.56}&\textcolor{blue}{95.84 ± 0.15}     \\  \hline  \hline
\end{tabular}%
}

\end{table}

Figure \ref{fig: training times} illustrates that the simplified models, when trained for extended periods, generally achieve higher accuracy and converge faster to their optimal values compared to fully trained models. 
Notably, when trained for a shorter duration (100 epochs), the simplified models not only outperform the fully trained counterparts by a larger margin but also require considerably fewer epochs to reach relatively high accuracies. Additionally, the accuracy gap between shorter and longer training durations is smaller for simplified models than for fully trained models.

To complement our empirical runtime measurements, we provide an asymptotic analysis of the computational cost. The complexity per layer is as follows:
\begin{itemize}
    \item \textbf{Trained GNN}: $O(|\mathbf{E}| \cdot d \cdot h + |\mathbf{V}| \cdot h^2)$, where the first term covers aggregation and the second term accounts for linear transformations.
    \item \textbf{UTMP}: $O(|\mathbf{E}| \cdot d)$ for aggregation only.
\end{itemize}
Here, $d$ is the input feature dimension and $h$ is the hidden dimension. The key saving comes from UTMP eliminating the $O(|\mathbf{V}| \cdot h^2)$ component related to learnable weight matrices, which is often the bottleneck in training GNNs on large graphs. This makes UTMP layers significantly more efficient, especially in resource-constrained settings. For simplified models (SMPNN), feature propagation can be precomputed once, reducing training to optimizing a single linear layer.

\section{HeaRT Split Results}
\label{sec; Heart split}
In this section, we show the results of the datasets split by the HeaRT evaluation setting \cite{li2023evaluating}.


\begin{table}[]
\centering
 \caption{Results on Cora, Citeseer, and Pubmed (\%) under HeaRT. }
 \begin{adjustbox}{width =1 \textwidth}
\begin{tabular}{cc|cccccc}

\toprule
 &\multirow{2}{*}{Models} & \multicolumn{2}{c}{Cora} &\multicolumn{2}{c}{Citeseer}  &\multicolumn{2}{c}{Pubmed}   \\ 

 & &MRR &Hits@10 &MRR &Hits@10 &MRR &Hits@10 \\
  \midrule
\multirow{5}{*}{Heuristic}&CN            & {9.78}  & {20.11} & {8.42}  &{18.68} & {2.28} & {4.78} \\
&AA            &{11.91} & {24.10}  & {10.82} & {22.20}  &{2.63} & {5.51} \\
&RA            &{11.81} & {24.48} & {10.84} &{22.86} & {2.47} & {4.9}  \\
&Shortest Path & {5.04} & {15.37} & {5.83}  & {16.26} &{0.86} &{0.38} \\
&Katz          & {11.41} &{22.77} &{11.19} & {24.84} & {3.01} &{5.98} \\

\midrule
\multirow{3}{*}{Embedding}&Node2Vec      & 14.47 ± 0.60               & 32.77 ± 1.29              & 21.17 ± 1.01              & 45.82 ± 2.01              & 3.94 ± 0.24              & 8.51 ± 0.77              \\
&MF            & 6.20 ± 1.42                & 15.26 ± 3.39              & 7.80 ± 0.79                & 16.72 ± 1.99              & 4.46 ± 0.32              & 9.42 ± 0.87              \\
&MLP           & 13.52 ± 0.65              & 31.01 ± 1.71              & 22.62 ± 0.55              & 48.02 ± 1.79              & 6.41 ± 0.25              &  15.04 ± 0.67             \\
\midrule
\multirow{4}{*}{GNN}&GCN           & {16.61 ± 0.30}               & {36.26 ± 1.14}              & 21.09 ± 0.88              & 47.23 ± 1.88              & 7.13 ± 0.27              & 15.22 ± 0.57             \\
&GAT           & 13.84 ± 0.68              & 32.89 ± 1.27              & 19.58 ± 0.84              & 45.30 ± 1.3                & 4.95 ± 0.14              & 9.99 ± 0.64              \\
&SAGE          & 14.74 ± 0.69              & 34.65 ± 1.47              & 21.09 ± 1.15              & 48.75 ± 1.85              & {9.40 ± 0.70}       & {20.54 ± 1.40}     \\
&GAE           & {18.32 ± 0.41}     & {37.95 ± 1.24}     & {25.25 ± 0.82}              & {49.65 ± 1.48}              & 5.27 ± 0.25              & 10.50 ± 0.46              \\
\midrule
\multirow{7}{*}{GNN+Pairwise Info}&SEAL          & 10.67 ± 3.46              & 24.27 ± 6.74              & 13.16 ± 1.66              & 27.37 ± 3.20               & 5.88 ± 0.53              & 12.47 ± 1.23             \\
&BUDDY         & 13.71 ± 0.59              & 30.40 ± 1.18              & 22.84 ± 0.36              & 48.35 ± 1.18              & 7.56 ± 0.18              & 16.78 ± 0.53             \\
&Neo-GNN       & 13.95 ± 0.39              & 31.27 ± 0.72              & 17.34 ± 0.84              & 41.74 ± 1.18              & {7.74 ± 0.30}              & {17.88 ± 0.71}             \\
&NCN           & 14.66 ± 0.95              & 35.14 ± 1.04              & {28.65 ± 1.21}     & {53.41 ± 1.46}              & 5.84 ± 0.22              & 13.22 ± 0.56             \\
&NCNC          & 14.98 ± 1.00               & {36.70 ± 1.57}               & {24.10 ± 0.65}               & {53.72 ± 0.97}     & {8.58 ± 0.59}              & {18.81 ± 1.16}      \\  
&NBFNet        & 13.56 ± 0.58              & 31.12 ± 0.75              & 14.29 ± 0.80               & 31.39 ± 1.34              & >24h                        & >24h                        \\
&PEG           & {15.73 ± 0.39}              & 36.03 ± 0.75              & 21.01 ± 0.77              & 45.56 ± 1.38              & 4.4 ± 0.41               & 8.70 ± 1.26               \\

\specialrule{.3em}{.2em}{.2em}

\multirow{4}{*}{GNN + GAE} 
& GCN &18.33 ± 0.32&37.29 ± 0.79&25.76 ± 0.68&50.44 ± 1.26&5.21 ± 0.27&10.55 ± 0.52  \\
& SAGE &14.34 ± 0.42&32.15 ± 1.32&21.35 ± 0.57&44.20 ± 0.98&4.12 ± 0.06&7.79 ± 0.24  \\
& GIN &12.81 ± 0.54&26.41 ± 1.21&16.29 ± 0.77&38.61 ± 1.01&3.64 ± 0.07&6.75 ± 0.11 \\
& GraphConv &12.48 ± 0.53&24.06 ± 1.30&13.36 ± 0.65&28.77 ± 1.14&3.38 ± 0.36&5.89 ± 0.88  \\ 
\hdashline

\multirow{3}{*}{S-GNN + GAE} 
& S-GCN &15.89 ± 0.26&34.27 ± 0.97&18.01 ± 2.59&43.78 ± 2.89&5.07 ± 0.24&9.95 ± 0.64  \\
& S-SAGE &13.59 ± 0.25&32.28 ± 0.84&14.45 ± 2.26&34.88 ± 3.38&3.04 ± 0.15&5.53 ± 0.52  \\
& S-GIN &12.87 ± 0.38&28.82 ± 1.30&12.89 ± 1.18&29.60 ± 2.78&3.78 ± 0.11&7.45 ± 0.15  \\ 
\hdashline

\multirow{3}{*}{NT-GNN + GAE} 
& NT-GCN &12.14&26.38&26.00&51.65&5.59&10.18  \\
& NT-SAGE &11.75&30.74&12.74&39.56&2.59&4.44  \\
& NT-GIN &0.77&0.19&1.16&1.10&1.89&1.94  \\
\midrule

\multirow{3}{*}{NCNC} 
& GCN &15.77 ± 0.59&36.51 ± 0.98&21.43 ± 1.12&49.56 ± 1.74&9.54 ± 0.59&21.84 ± 1.00  \\
& SAGE &12.18 ± 0.57&30.34 ± 1.25&16.22 ± 1.72&31.34 ± 4.09&6.48 ± 0.37&14.84 ± 1.10  \\
& GIN &12.23 ± 0.48&30.76 ± 1.17&16.77 ± 1.05&33.43 ± 2.64&6.54 ± 0.27&14.62 ± 0.46  \\
\hdashline

\multirow{3}{*}{NCNC} 
& SGCN &15.92 ± 0.80&38.04 ± 0.63&23.55 ± 0.82&53.72 ± 1.34&8.29 ± 0.65&18.28 ± 1.49  \\
& SSAGE &11.92 ± 1.35&29.96 ± 0.93&20.04 ± 3.61&42.90 ± 6.60&4.87 ± 0.66&10.41 ± 1.57  \\
& SGIN &12.59 ± 0.75&30.06 ± 2.22&22.35 ± 0.83&45.87 ± 1.72&4.95 ± 0.30&10.38 ± 0.70  \\
    
 \bottomrule
\end{tabular}
 \label{table:small_newsetting}
 \end{adjustbox}
\end{table}


\begin{table}[]
\centering
 \caption{Results on OGB datasets (\%) under HeaRT.}
 \begin{adjustbox}{width =1 \textwidth}
\begin{tabular}{c|cccccccc}

\toprule
 \multirow{2}{*}{Models} & \multicolumn{2}{c}{ogbl-collab} &\multicolumn{2}{c}{ogbl-ddi}  &\multicolumn{2}{c}{ogbl-ppa} &    \multicolumn{2}{c}{ogbl-citation2}  \\ 
 
  &MRR &Hits@20 &MRR &Hits@20 &MRR &Hits@20 &MRR & Hits@20\\
 \midrule

CN & {4.20} & {16.46} & 6.71 &38.69 & 25.70  &68.25 & 17.11 &41.73 \\
AA & {5.07} & {19.59} &{6.97} & {39.75} &{26.85} &{70.22} &{17.83}  & {43.12} \\
RA & {6.29} & {24.29} &{8.70} & {44.01}       &{28.34} &{71.50}  & 17.79	&43.34 \\
Shortest Path & {2.66} & {15.98} &{0} & {0}  &{0.54}  & {1.31}  & >24h & >24h \\
Katz &{6.31} & {{24.34}} & 6.71&38.69 & 25.70&68.25 & 14.10 & 35.55 \\ \midrule
Node2Vec & 4.68 ± 0.08 & 16.84 ± 0.17 & 11.14 ± 0.95 & 63.63 ± 2.05 & 18.33 ± 0.10  & 53.42 ± 0.11 & 14.67 ± 0.18 & 42.68 ± 0.20 \\
MF & 4.89 ± 0.25 & 18.86 ± 0.40 & {13.99 ± 0.47} & 59.50 ± 1.68 & 22.47 ± 1.53 & 70.71 ± 4.82  & 8.72 ± 2.60 & 29.64 ± 7.30 \\
MLP & 5.37 ± 0.14 & 16.15 ± 0.27 & N/A & N/A  & 0.98 ± 0.00 & 1.47 ± 0.00& 16.32 ± 0.07 & 43.15 ± 0.10  \\ \midrule
GCN & {6.09 ± 0.38} &{22.48 ± 0.81} & {13.46 ± 0.34} & 64.76 ± 1.45 & 26.94 ± 0.48 & 68.38 ± 0.73 & 19.98 ± 0.35 & {51.72 ± 0.46} \\
GAT & 4.18 ± 0.33 & 18.30 ± 1.42 & {12.92 ± 0.39} & {66.83 ± 2.23}  & OOM & OOM  & OOM  & OOM     \\
SAGE & 5.53 ± 0.5 & 21.26 ± 1.32 & 12.60 ± 0.72 & {67.19 ± 1.18} & 27.27 ± 0.30 & 69.49 ± 0.43  & {22.05 ± 0.12} & {53.13 ± 0.15}\\
GAE & {OOM} &{OOM}&{3.49 ± 1.73} & {17.81 ± 9.80} & OOM & OOM  & OOM & OOM            \\ \midrule
SEAL & {6.43 ± 0.32} & 21.57 ± 0.38 & 9.99 ± 0.90 & 49.74 ± 2.39 & {29.71 ± 0.71}& {76.77 ± 0.94} & {20.60 ± 1.28} & 48.62 ± 1.93 \\
BUDDY & 5.67 ± 0.36 & {23.35 ± 0.73}  & 12.43 ± 0.50 & 58.71 ± 1.63  & 27.70 ± 0.33 & 71.50 ± 0.68 & 19.17 ± 0.20  & 47.81 ± 0.37   \\
Neo-GNN & 5.23 ± 0.9 & 21.03 ± 3.39 & 10.86 ± 2.16 & 51.94 ± 10.33 & 21.68 ± 1.14 & 64.81 ± 2.26  & 16.12 ± 0.25   & 43.17 ± 0.53        \\
NCN & 5.09 ± 0.38 & 20.84 ± 1.31 & 12.86 ± 0.78 & {65.82 ± 2.66} & {35.06 ± 0.26} & {81.89 ± 0.31}  & {23.35 ± 0.28}& {53.76 ± 0.20}      \\
NCNC & 4.73 ± 0.86 & 20.49 ± 3.97  & >24h    & >24h & {33.52 ± 0.26}   & {82.24 ± 0.40}  & 19.61 ± 0.54  & 51.69 ± 1.48       \\ 
NBFNet & OOM  & OOM     & >24h   & >24h      & OOM        & OOM     & OOM    & OOM                       \\
PEG & 4.83 ± 0.21 & 18.29 ± 1.06 & 12.05 ± 1.14  & 50.12 ± 6.55  & OOM   & OOM    & OOM     & OOM                       \\
\specialrule{.3em}{.2em}{.2em}
GCN + GAE &OOM&OOM&6.52 ± 0.51&34.48 ± 1.52&OOM&OOM&OOM&OOM\\
SAGE + GAE &OOM&OOM&5.33 ± 0.21&29.30 ± 1.44&OOM&OOM&OOM&OOM \\
GIN + GAE &OOM&OOM&12.15 ± 0.26&54.21 ± 0.95&OOM&OOM&OOM&OOM \\
GraphConv + GAE &OOM&OOM&13.60 ± 0.26&56.36 ± 0.45&OOM&OOM&OOM&OOM \\
\hdashline
SGCN + GAE &OOM&OOM&3.82 ± 1.97&18.21 ± 10.38&OOM&OOM&OOM&OOM \\
SSAGE + GAE &OOM&OOM&2.71 ± 0.22&10.98 ± 0.76&OOM&OOM&OOM&OOM \\
SGIN + GAE &OOM&OOM&0.40&0.00&OOM&OOM&OOM&OOM \\
\hdashline
NTGCN + GAE &1.07&0.66&5.65&33.23&19.58&30.60&0.40&0.00 \\
NTSAGE + GAE &0.40&0.00&0.60&0.00&1.96&3.26&0.40&0.00 \\
NTGIN + GAE &0.40&0.00&0.40 &0.00&0.40&0.00&0.40&0.00 \\
\midrule
NCNC + GCN &3.53 ± 0.29&13.99 ± 1.44&OOM&OOM&35.75 ± 1.23&75.94 ± 1.51&21.52 ± 0.71&53.08 ± 1.61 \\
NCNC + SAGE &2.97 ± 0.63&11.28 ± 2.77&OOM&OOM&35.37 ± 1.09&71.62 ± 1.70&20.73 ± 0.63&51.87 ± 0.86 \\
NCNC + GIN &2.93 ± 0.64&11.28 ± 2.56&OOM&OOM&18.81 ± 6.81&38.51 ± 9.97&21.36 ± 0.89&52.48 ± 1.09 \\
\hdashline
NCNC + SGCN &3.60 ± 0.87&14.06 ± 3.79&OOM&OOM&33.98 ± 1.90&70.87 ± 3.55&20.25 ± 0.38&50.95 ± 0.33 \\
NCNC + SSAGE &3.79 ± 0.54&15.17 ± 2.15&OOM&OOM&34.09 ± 0.55&69.77 ± 1.57&20.55 ± 0.29&51.08 ± 0.39 \\
NCNC + SGIN &2.38 ± 0.25&6.72 ± 0.85&OOM&OOM&20.46 ± 1.41&40.16 ± 1.90&20.97 ± 0.86&51.13 ± 1.70 \\
 \bottomrule
\end{tabular}
 \label{table:ogb_newsetting}
 \end{adjustbox}

\end{table}

\end{document}